\documentclass{article}

\usepackage[final, nonatbib]{neurips_2023}

\usepackage[utf8]{inputenc} % allow utf-8 input
\usepackage[T1]{fontenc}    % use 8-bit T1 fonts
\usepackage{hyperref}       % hyperlinks
\usepackage{url}            % simple URL 
\usepackage{booktabs}       % professional-quality tables
\usepackage{textpos}  %
\usepackage{amsfonts}       % blackboard math symbols
\usepackage{nicefrac}       % compact symbols for 1/2, etc.
\usepackage{microtype}      % microtypography
\usepackage{xcolor}         % colors
\usepackage{amsmath}
\usepackage{amssymb}
\usepackage{makecell}
\usepackage[export]{adjustbox}
\usepackage[normalem]{ulem}
\usepackage{xspace, eucal}
\usepackage{tabularx}
\usepackage{bm}
\usepackage{mathtools,amssymb}
\usepackage{tabulary,multirow,overpic,xcolor,subfloat}
\usepackage{float}
\usepackage{subcaption}
\usepackage{multirow}
\usepackage{multicol}

\usepackage{pgfplots}

\definecolor{citecolor}{RGB}{34,139,34}
\definecolor{linkcolor}{HTML}{ED1C24}

\newcommand{\app}{\raise.17ex\hbox{$\scriptstyle\sim$}}

\newcolumntype{x}[1]{>{\centering\arraybackslash}p{#1pt}}
\newcolumntype{y}[1]{>{\raggedright\arraybackslash}p{#1pt}}

\newlength\savewidth\newcommand\shline{\noalign{\global\savewidth\arrayrulewidth
  \global\arrayrulewidth 1pt}\hline\noalign{\global\arrayrulewidth\savewidth}}

\newcommand{\defeq}{\coloneqq}
\newcommand{\tablestyle}[2]{\setlength{\tabcolsep}{#1}\renewcommand{\arraystretch}{#2}\centering\footnotesize}

\makeatletter\renewcommand\paragraph{\@startsection{paragraph}{4}{\z@}
  {.5em \@plus1ex \@minus.2ex}{-.5em}{\normalfont\normalsize\bfseries}}\makeatother

\definecolor{Gray}{gray}{0.5}

\def\eg{\emph{e.g.}}

\def\ie{\emph{i.e.}}

\newcommand{\modelname}{\textup{DatasetDM}\xspace}

\title{\modelname: Synthesizing Data with Perception Annotations Using Diffusion Models}

\author{%
  Weijia Wu\textsuperscript{$1,$}\textsuperscript{$3$} 
  \quad Yuzhong Zhao\textsuperscript{$2$}
  \quad Hao Chen\textsuperscript{$1$}
  \quad Yuchao Gu\textsuperscript{$3$}
  \quad Rui Zhao\textsuperscript{$3$}
  \quad Yefei He\textsuperscript{$1$} 
  \\
  \quad \textbf{Hong Zhou\textsuperscript{$1$}$^{*}$}
  \quad \textbf{Mike Zheng Shou\textsuperscript{$3$}\thanks{H. Zhou and M. Shou are the corresponding authors.}}
  \quad \textbf{Chunhua Shen\textsuperscript{$1$,$4$}}
  \\[0.2cm]
  \textsuperscript{1}Zhejiang University, China 
  ~~~
  \textsuperscript{2}University of Chinese Academy of Sciences, China
  \\
  \textsuperscript{3}National University of Singapore
  ~~~
  \textsuperscript{4}Ant Group
}

\begin{document}

\maketitle

\begin{abstract}
  Current deep networks are 
  %extremely
  very 
  data-hungry and benefit from training on large-scale datasets, which are often time-consuming to collect and annotate.
  By contrast, synthetic data can be generated infinitely 
  %from
  using 
  generative models such as DALL-E and %Stable
  diffusion models, with minimal effort and cost.
  In this paper, we present \modelname, a 
  %generalized
  generic 
  dataset generation model that can produce 
  %an unlimited number of 
  diverse 
  synthetic images and the corresponding high-quality perception annotations %, such as 
  (\eg, segmentation masks, and depth).
  Our method builds upon the pre-trained diffusion model and extends text-guided image synthesis to perception data generation. 
  We show that the rich latent code of the diffusion model can be effectively decoded as accurate perception annotations using a decoder module.
  Training the decoder only needs less than %\textbf
  {$\bf 1\%$} (around 100 images) %existing
  manually 
  labeled images, enabling the generation of an infinitely large annotated dataset.
  Then these synthetic data can be used for training 
  %any 
  various 
  perception models for downstream tasks.

  To showcase the power of the proposed approach, we generate datasets with rich dense pixel-wise labels for a wide range of downstream tasks, including semantic segmentation, instance segmentation, and depth estimation.
   Notably, it achieves (1) state-of-the-art results on semantic segmentation and instance segmentation;
   (2) significantly more robust on domain generalization than using the real data alone; and state-of-the-art results in zero-shot segmentation setting; and 
   (3) flexibility for efficient application and novel task composition (\eg, image editing).
   
   The project website is at:  
   \href{https://weijiawu.github.io/DatasetDM_page/}{\color{blue}{\tt https://weijiawu.github.io/DatasetDM}}.
   % and \href{https://github.com/showlab/DatasetDM}{\color{blue}{$\tt DatasetDM$}}, respectively.
\end{abstract}

\section{Introduction}
Modern deep-learning models %used
 for perception tasks often require a large amount of labeled data to achieve good performance.
Unfortunately, collecting large-scale data and labeling the corresponding pixel-level annotations is a time-consuming and expensive process.
For example, collecting images of urban driving scenes requires physical car infrastructure, and labeling a segmentation annotation for a single urban image in Cityscapes~\cite{cordts2016cityscapes} can take up to 60 minutes.
Moreover, in certain specialized domains, such as medical or human facial data, collecting relevant information can be challenging or even impossible, owing to privacy concerns or other factors.
The above challenges can be a barrier to advancing artificial intelligence in computer vision.

\begin{figure}[t]
    \centering
    \includegraphics[width=\linewidth]{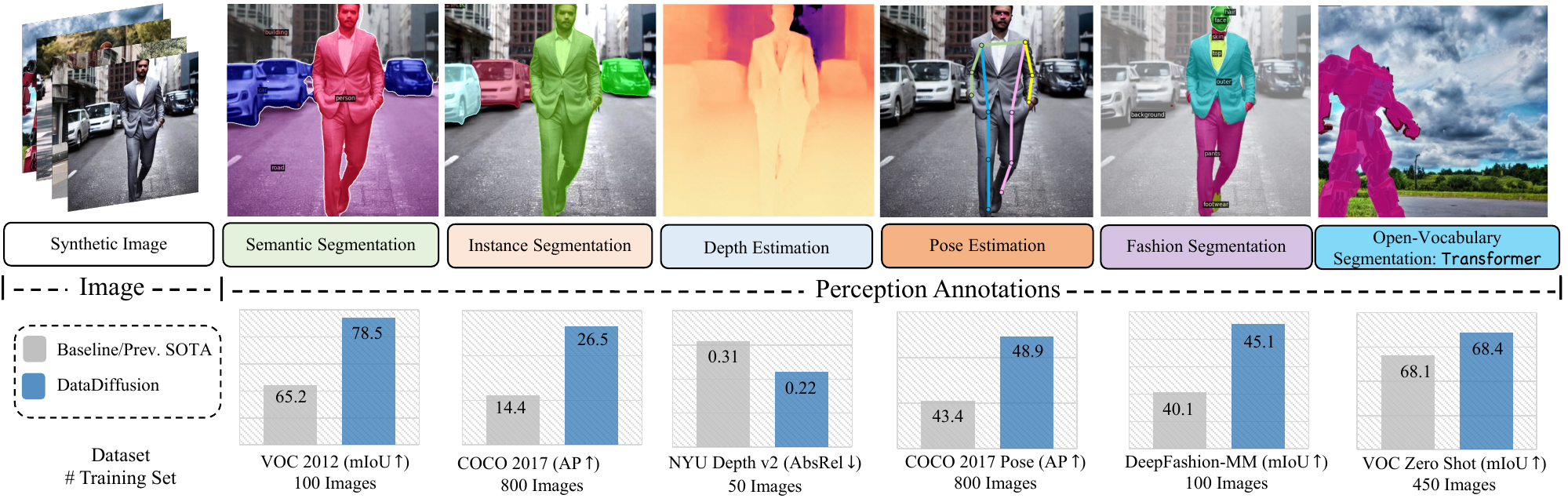}
  %  %%%\vspace{-3mm}
    \caption{
        \textbf{Synthetic Data from \modelname}. The high-quality and infinitely numerous synthetic images with perception annotation can yield significant improvements for various downstream tasks.
        }
    \label{fig:teaser}
   \vspace{-1mm}
\end{figure}

To reduce costs, many previous researchers have primarily focused on weakly supervised~\cite{zhang2020survey} and unsupervised solutions~\cite{wilson2020survey} to address the problem. 
For instance, certain segmentation priors~\cite{ahn2018learning,akiva2021towards,lin2016scribblesup} use weak or inexpensive labels to train robust segmentation models.
With the advancement of generative models, such as DALL-E~\cite{ramesh2022hierarchical}, Stable Diffusion~\cite{rombach2022high}, some researchers have begun to explore the potential of synthetic data, attempting to use it to assist in training models, or even replace real data for perception task.
Most works focus on classification, face recognition~\cite{lin2022raregan,wang2022deep,he2022synthetic}, salient object detection~\cite{wu2022synthetic} and segmentation tasks~\cite{li2023guiding,wu2023diffumask,zhang2021datasetgan}, with only a minority trying to address problems such as human pose estimation~\cite{ge2020pose} or medical image analysis~\cite{han2023medgen3d}.
In the era of GANs, DatasetGAN~\cite{zhang2021datasetgan} and BigDatasetGAN~\cite{li2022bigdatasetgan} are recognized as pioneering works that utilize the feature space of pre-trained GANs and design a shallow decoder for generating pixel-level labels in the context of segmentation tasks.
Following the two works, Hands-off~\cite{xu2022handsoff} extends this approach to multitasking scenarios, such as depth estimation.
However, these methods still suffer from three major drawbacks: 
% Due to the limitations of the representation ability of early (up to 2020) GAN models.
1) Due to the limitations of the representation ability of early (up to 2020) GAN models, the quality of the synthesized data is often dissatisfactory, leading to an inferior performance on downstream tasks.
2) These methods primarily focus on independent downstream tasks and no one tries to explore a unified data synthesis paradigm with a generalized decoding framework.
3) The training cost is still relatively high, while these methods do not make full use of the visual knowledge contained within the latent codes of powerful text-to-image models.

By leveraging large-scale datasets of image-text pairs (\eg{}, LAION5B~\cite{schuhmann2022laion}), recent text-to-image diffusion models~(\eg{}, Stable Diffusion~\cite{rombach2022high}) present phenomenal power in generating diverse and high-fidelity images with rich texture, diverse content, and reasonable structures.
The phenomenon suggests that large text-to-image diffusion models can \textit{implicitly } learn valuable, and rich high-level and low-level visual representations from massive image-text pairs.
It is natural %and seamless to think:
ask:  \textit{Can we leverage the knowledge learned by these models to generate perception annotations and extend the paradigm of text-to-image generation to text-to-data generation?}

In this paper, built upon the powerful text-to-image diffusion model, we present \modelname, a generalized dataset generation model that can produce an unlimited number of synthetic images and the corresponding perception annotations, as shown in Fig. \ref{fig:teaser}.
The key to our approach is a unified perception decoder, namely P-Decoder, that decodes the latent code of the diffusion model to perception annotations.
Inspired by align/instruct tuning from LLM~\cite{wang2022self}, a method for inducing output following capability with minimal human-labeled,
we only use less than { $\bf 1\%$} manually labeled images to train the decoder, enabling infinite annotated data generation. 
The generated data can subsequently be utilized to train any perception models for various downstream tasks, including but not limited to segmentation, depth estimation, and pose estimation.
To maintain the robust image generation capabilities of the diffusion model, we freeze the model weights and %utilize
use image inversion to extract the latent code, \ie{}, multi-scale features, which are then fed into the unified P-Decoder. 
The P-Decoder accommodates various downstream tasks within a unified transformer framework, with only minor variations as depicted in Fig. \ref{fig:decorder}.

To summarize, our contributions are four-fold:
\begin{itemize}
    \item
    We introduce \modelname: a versatile dataset generation model featuring a perception decoder capable of producing an unlimited quantity of high-fidelity synthetic images, along with various perception annotations, including depth, segmentation, and human pose estimation.

    \item 
    Visual align/instruct tuning, a method for inducing output following capability with minimal human-labeled data.
    With less than $1\%$ of the original dataset, \ie{}, around 100 images, \modelname pushes the limits of text-to-image generation, pioneering a novel paradigm: \textit{text-to-data generation}. 
    This breakthrough is made possible by leveraging the rich latent code of the pre-trained diffusion model.
    
    \item Experiments demonstrate that the existing perception models trained on synthetic data generated by \modelname exhibit outstanding performance on \textbf{six} datasets across \textbf{five} different downstream tasks. 
    For instance, the synthetic data delivers remarkable gains of $13.3\%$ mIoU and $12.1\%$ AP for semantic segmentation on VOC 2012 and instance segmentation on COCO 2017, respectively.
    
    \item 
    Text-guided data generation allows for the generation of a diverse range of data, which has been shown to provide a more robust solution for domain generalization and long-tail data distribution. 
    Moreover, \modelname offers a flexible approach for novel task composition, as exemplified by its ability to facilitate image editing (see Fig. \ref{fig:Examples1}).

\end{itemize}

\section{Related Work}
\label{sec:related_work}

%\subsection
{\bf Text-to-Image Generation.}
Several mainstream methods exist for the task, including Generative Adversarial Networks (GANs)\cite{goodfellow2020generative}, Variational Autoencoders (VAEs)\cite{kingma2013auto}, flow-based models~\cite{dinh2014nice}, and Diffusion Probabilistic Models (DPMs)~\cite{sohl2015deep,rombach2022high,ho2020denoising,gu2023mix}.
Recently, as a likelihood-based method, diffusion models have gained significant attention with promising generation abilities.
They match the underlying data distribution by learning to reverse a noising process.
Thanks to the high-quality image synthesis and the stability of training, diffusion models are quickly emerging as a new frontier~\cite{rombach2022high,ramesh2022hierarchical,ho2022cascaded,saharia2022photorealistic} in the field of image synthesis.
Text-guided diffusion models are used for text-conditional image generation, where a text prompt $\mathcal{P}$ guides the generation of content-related images $\mathcal{I}$ from a random Gaussian noise $z$. Visual and textual embeddings are typically fused using cross-attention.
Recent large text-to-image diffusion models, such as Stable Diffusion~\cite{rombach2022high} of Stability AI, DALL-E2~\cite{ramesh2022hierarchical} of OpenAI, and Imagen~\cite{saharia2022photorealistic} of Google, have shown powerful performance in generating diverse and high-fidelity images with rich textures and reasonable structures. 
Their impressive synthesis capabilities suggest that these models can implicitly learn valuable representations with the different semantic granularity from large-scale image-text pairs.
In this paper, we leverage these learned representations~(latent codes), to extend the paradigm of text-to-image generation to text-to-data generation.

%\subsection
{\bf Synthetic Datasets for Perception Task.}
Previous studies in dataset synthesis, such as Virtual KITTI~\cite{gaidon2016virtual} and Virtual KITTI 2~\cite{cabon2020virtual}, primarily rely on 3D computer simulations to address standard 2D computer vision problems, such as object detection~\cite{pepik2012teaching}, scene understanding~\cite{satkin2012data}, and optical flow estimation~\cite{butler2012naturalistic}. 
However, these methods are limited by the domain of 3D models and cannot be generalized to arbitrary scenes and open-set categories.
For instance, Virtual KITTI is exclusively focused on autonomous driving scenes and supports only 20 commonly occurring categories, which cannot be extended to open-scene domains like the COCO benchmark~\cite{lin2014microsoft}.

In contrast, synthetic data generated using generation models~(\ie{}, GAN~\cite{goodfellow2020generative,ling2021editgan,xu2022handsoff} and Diffusion Model~\cite{sohl2015deep}) are more flexible and can support a wider range of tasks and open-world scenes for various downstream tasks, such as classification task~\cite{he2022synthetic}, face recognition~\cite{he2022synthetic}, salient object detection~\cite{wu2022synthetic}, semantic segmentation~\cite{li2022bigdatasetgan,zhang2021datasetgan,baranchuk2021label,zhao2023generative}, and human pose~\cite{ge2020pose}.
Inspired by the success of large-scale generative models, such as Stable Diffusion~\cite{rombach2022high}, trained on massive datasets like LAION5B~\cite{schuhmann2022laion}, recent studies have begun to explore the potential of powerful pre-trained diffusion generative models.
%
% Handsoff~\cite{xu2022handsoff} explores the use of GANs to generate semantic and depth annotations.
%
Based on DDPM~\cite{ho2020denoising}, DatasetDDPM~\cite{baranchuk2021label} design several CNN layers as the annotation decoder to generate semantic segmentation data by performing a multi-class task.
Li \textit{et al.}~\cite{li2023guiding} utilized Stable Diffusion and Mask R-CNN pre-trained on the COCO dataset~\cite{lin2014microsoft} to design and train a grounding module for generating images and semantic segmentation masks.
DiffuMask~\cite{wu2023diffumask} produces synthetic image and semantic mask annotation by exploiting the potential of the cross-attention map between text and image from the text-supervised pre-trained Stable Diffusion model.
The above methods focus on semantic segmentation, which cannot handle tasks such as instance segmentation. 
In this paper, we take a further step by utilizing a generalized perception decoder to parse the rich latent space of the pre-trained diffusion model, enabling the generation of perception for a variety of downstream tasks, including depth, segmentation, and human pose.

{\bf Diffusion Model for Perception Task.}
Some recent works~\cite{zhao2023unleashing,xu2023open,wu2022medsegdiff} has also attempted to directly employ diffusion models for perceptual tasks.
VPD~\cite{zhao2023unleashing} explores the direct application of pre-trained Stable Diffusion to design perception models.
ODISE~\cite{xu2023open} unify pre-trained text-image diffusion and discriminative models to perform open-vocabulary panoptic segmentation.
Different from these approaches, our focus lies in synthetic data augmentation for perception tasks, and design a unified transformer-based decoder to enhance more perception tasks from data aspect.

\section{Methodology}
\label{sec:methodology}

\subsection{Formulation}
Given a language prompt $\mathcal{S}$, text-guided diffusion models generate content-related images $\mathcal{I} \in \mathcal{R}^{H \times W \times 3}$ from a random Gaussian noise $\bm{z} \sim \mathcal{N}(\mathbf{0},\mathbf{I})$. 
The standard text-guided image denoising processing can be formulated as: $\mathcal{I} = \Phi_{\text{T2I}}(\bm{z}, \mathcal{S})$, where $\Phi_{\text{T2I}}(\cdot)$ refers to a pre-trained text-to-image diffusion model.
In this paper, we adopt Stable Diffusion~\cite{rombach2022high} as the base for the diffusion model, which consists of three components: 
a text encoder $\tau_\theta(\cdot)$ for embedding prompt $\mathcal{S}$; 
% VAGAN
a pre-trained variational autoencoder~(VAE)~\cite{esser2021taming} that encodes $\mathcal{E}(\cdot)$ and decodes $\mathcal{D}(\cdot)$ latent code of images; 
and a time-conditional UNet~($\epsilon_{\theta}(\cdot)$)~\cite{ronneberger2015u} that gradually denoises the latent vectors.
To fuse visual and textual embeddings, cross-attention layers are typically used in the UNet for each denoising step. 
The denoising process is modeled as a Markov chain: $\bm{x}_{t-1} = f(\bm{x}_t, \epsilon_{\theta})$,
where $\bm{x}_{t}$ denote latent code at timestep $t$, and $t\in[1,T]$.
The latent noise at the final timestep $T$, denoted as $\bm{x}_{T}$, is equivalent to the random Gaussian noise $\bm{z}$.
$f(\cdot)$ %refer to
is 
the denoising function~\cite{ho2020denoising}.

In this paper, we design a perception decoder that can effectively parse the latent space of the UNet $\epsilon_{\theta}(\bm{x}_t, t, \tau_\theta(\mathcal{S}))$.
By doing so, we extend the \textit{text-to-image} generation approach to a 
\textit{text-to-data} paradigm:
\begin{align}
  \{\mathcal{I},\mathcal{P}_{1:k}\} = \Phi_{\text{T2D}}(\bm{z}, \mathcal{S}),
\end{align}
where $\mathcal{P}_{1:k}$ denotes the corresponding perception annotations, and $k$ is the number of the supported downstream tasks.
In fact, the paradigm can support any image-level perception task, such as semantic segmentation, instance segmentation, pose estimation, and depth estimation.

\subsection{Method Overview}
This paper introduces a novel paradigm called \textit{text-to-data generation}, which extends text-guided diffusion models trained on large-scale image-text pairs.
Our key insight is that using a small amount of real data (using less than $1\%$ existing labeled dataset) and a 
%generalized
generic 
perception decoder to interpret the diffusion latent spaces, results in the generation of infinite and diverse annotated data.
Then the synthetic data can be used to train any existing perception methods and apply them to real images.

The proposed \modelname framework, presented in Fig. \ref{fig:pipeline}, comprises two stages. 1) The first stage---\textbf{Training}---involves using diffusion inversion (%Sec.%
\S\ref{sec:Inversion}) to obtain the latent code of the real image and extract the text-image representations (%Sec. 
\S\ref{sec:Representation}). These representations and their corresponding annotations are then used to train the perception decoder (\S\ref{sec:Decoder}). 
2) The second stage---\textbf{Inference} (\S\ref{sec:Prompting})---%
%, 
%utilizes
uses 
GPT-4 to enhance the diversity of data and generates abundant images, while the P-Decoder produces corresponding perception annotations such as masks and depth maps.

\subsection{Hypercolumn Representation Extraction}
\begin{figure}
    \centering
    \includegraphics[width=\linewidth]{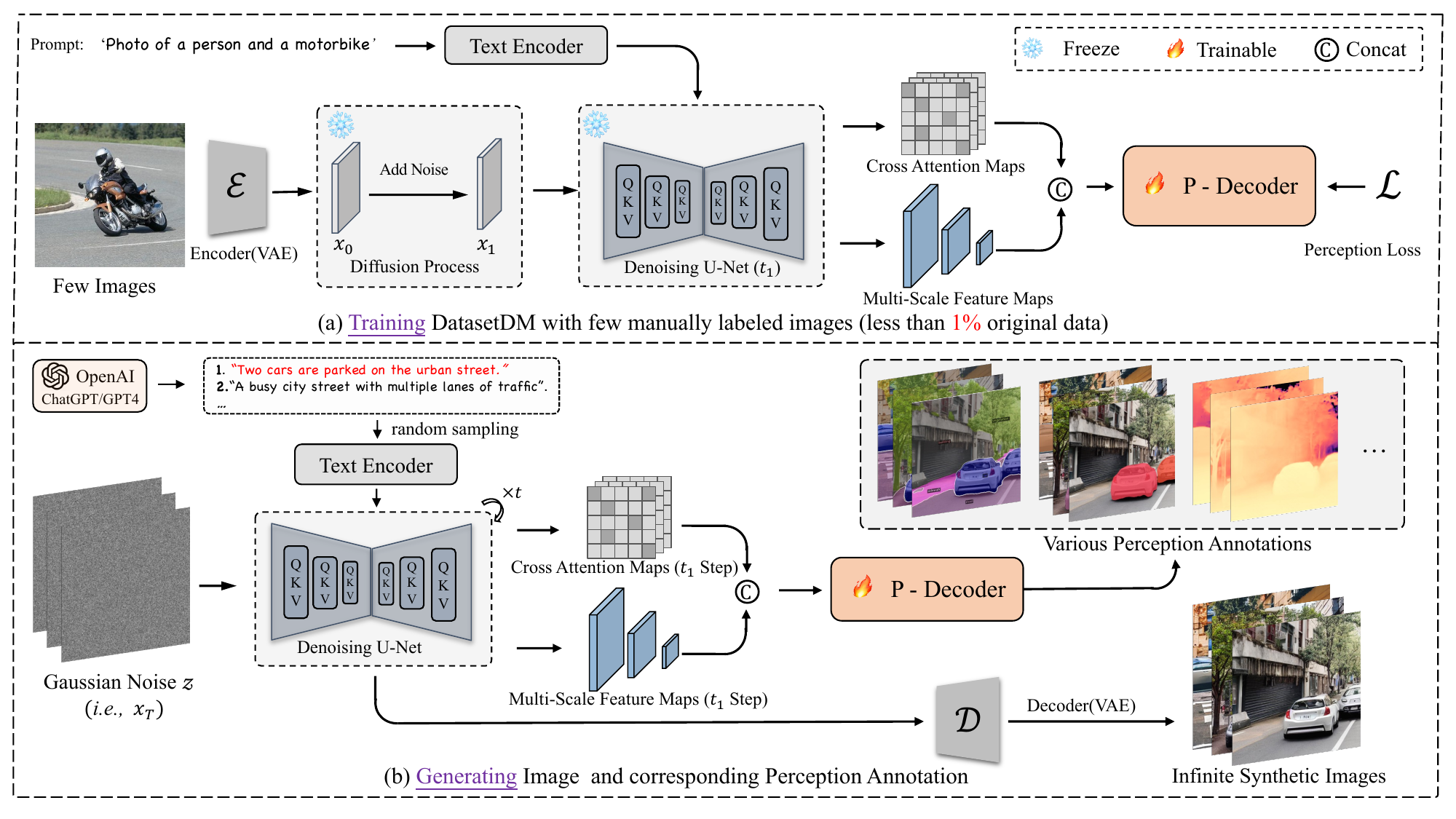}
  %  %%%\vspace{-4mm}
    \caption{
        \textbf{The 
        %whole
        overall 
        framework of \modelname.}
        \modelname consists of two main steps: 1) Training. Using
        diffusion inversion 
        to extract the latent code from a small amount of data and then train the perception decoder.
        2) Text-guided data generation. A large language model such as GPT-4 is utilized to prompt infinite and diverse data generation for various downstream tasks.
        }
    \label{fig:pipeline}
    \vspace{-2mm}
\end{figure}

The first step in the training stage of \modelname is to extract the hypercolumn representation of real images from the latent space of the diffusion model, as shown in Fig. \ref{fig:pipeline}(a).
To achieve this, we employ the diffusion inversion technique, which involves adding a certain level of Gaussian noise to the real image and then extracting the features from the UNet during the denoising process.

\textbf{Image Inversion for Diffusion Model.}
\label{sec:Inversion}
Give a real image $\mathcal{X} \in \mathcal{R}^{H \times W \times 3}$ from 
the 
training set, the diffusion inversion~(diffusion forward process) is a process 
that 
approximates the posterior distribution $q(\bm{x}_{1:T}|\bm{x}_0)$, where $\bm{x}_0=\mathcal{E}(\mathcal{X})$. 
This process is not trainable and is based on a fixed Markov chain that gradually adds Gaussian noise to the image, following a pre-defined noise schedule $\beta_1, \dotsc, \beta_T$~\cite{ho2020denoising}:
\begin{align} 
q(\bm{x}_t|\bm{x}_{t-1}) \defeq \mathcal{N}(\bm{x}_t;\sqrt{1-\beta_t}\bm{x}_{t-1},\beta_t \mathbf{I}), \label{eq:forwardprocess}
\end{align}
where $t$ represents the $t$-th time step, and we set it to $1$ for a single forward pass in our paper. 
A single forward pass for visual representation extraction usually provides two advantages, \ie, faster convergence 
%with fewer training time 
and better performance \cite{xu2023open}.

\textbf{Text-Image Representation Fusion.}
\label{sec:Representation}
With the latent code $\bm{x}_{t}$ of the real image and the corresponding language prompt $\mathcal{S}$, we extract the multi-scale feature maps and cross attention maps from the UNet $\epsilon_{\theta}$ as follows:
\begin{align}
  \{\mathcal{F},\mathcal{A}\} = \epsilon_{\theta}(\bm{x}_t, t, \tau_\theta(\mathcal{S})),
\end{align}
where $\mathcal{S}$ for training set is simply defined using a template ``\texttt{a photo
of a [CLS]$_1$, [CLS]$_2$, \dots}''.
During the data generation phase, GPT-4 is used to provide diverse prompt languages. 
The variable $\mathcal{F}$ denotes the multi-scale feature maps from four layers of the U-Net, corresponding to four different resolutions, \ie{}, $8\times8$, $16\times16$, $32\times32$, and $64\times64$, as illustrated in Fig. \ref{fig:pipeline}.
Additionally, $\mathcal{A}$ represents the cross-attention maps of text-to-image from the $16$ cross-attention layers in the U-Net, which implement the function $\mathcal{A} = \text{softmax}\left(\frac{QK^T}{\sqrt{d}}\right)$, where $d$ is the latent projection dimension. 
We group the $16$ cross-attention maps into $4$ groups with the same resolutions, and compute their average within each group, which results in the average cross-attention maps $\mathcal{\hat{A}}$.

Prior works \cite{wu2023diffumask,zhao2023unleashing,hertz2022prompt} have proved the effectiveness of class-discriminative and localization of cross-attention map between the visual embedding and the conditioning text features.
Thus we concatenate the cross-attention maps $\mathcal{\hat{A}}$ %with
and 
the multi-scale feature maps $\mathcal{F}$ %as
to obtain 
the final extracted hyper-column representation, and further use a
$1\times1$ convolution to fuse them: $\mathcal{\hat{F}}=\text{Conv}
([\mathcal{F},\mathcal{\hat{A}}])$.

\subsection{Perception Decoder}
\label{sec:Decoder}
\begin{figure}
    \centering
    \includegraphics[width=\linewidth]{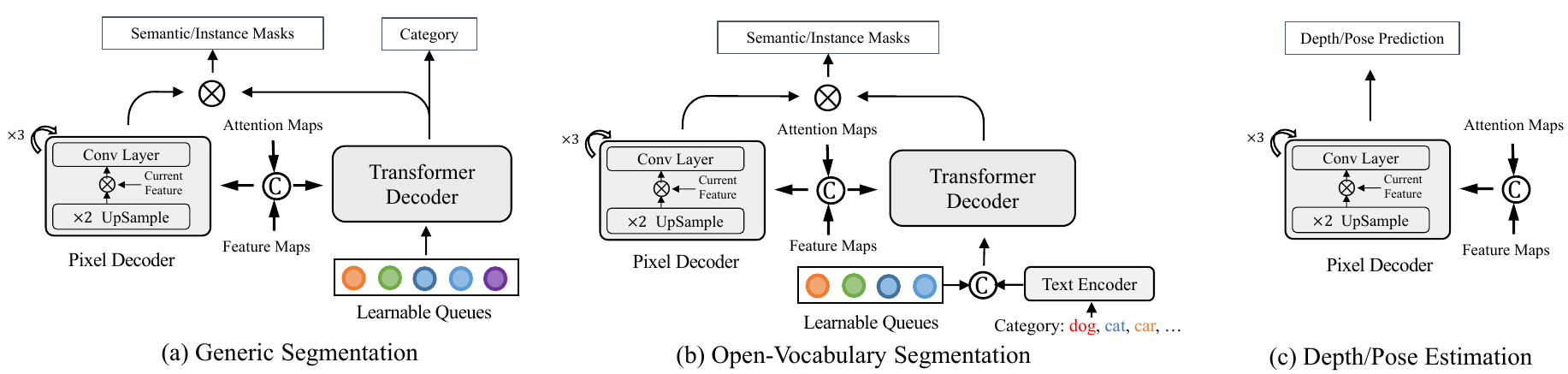}
  %  %%%\vspace{-4mm}
    \caption{
        \textbf{
        %Different
    Various 
        types
        of tasks with our proposed P-Decoder. 
        } The proposed decoder is a generalized architecture for the six supported tasks, with only minor variations required for different downstream applications, \ie{}, determining whether to activate certain layers. 
        }
    \label{fig:decorder}
    %%%\vspace{-2mm}
\end{figure}

The P-Decoder is utilized to translate the representation $\mathcal{\hat{F}}$ into perception annotations, which are not limited to a specific type for each downstream task. 
To achieve this goal, inspired by previous works~\cite{cheng2022masked,zou2022generalized}, we devised a generalized architecture.
This architecture is depicted in Fig. \ref{fig:decorder}, with only \textit{minor variations} (\ie{}, whether to startup some layers) for each downstream task.
For example, the pixel decoder and transformer decoder are required for generic segmentation, while only the pixel decoder is necessary for depth and pose estimation.

% 1) Transformer-based~\cite{cheng2022masked} for the class/instance aware tasks, \ie{} semantic and instance segmentation, as shown in Fig. \ref{fig:decorder}(a-b). 2) Simple multiple upsampling CNN layers for the tasks that the format of the output is pre-defined, \ie{} no class/instance distinction, such as depth and pose estimation in Fig. \ref{fig:decorder}(c-d).

\textbf{Generic Image Segmentation.} In Fig. \ref{fig:decorder}-(a), we present the adaptation for semantic and instance segmentation tasks, which includes two components: the pixel decoder and the transformer decoder.
Similar to Mask2former~\cite{cheng2022masked}, the transformer decoder comprises a stack of transformer layers with cross-attention and self-attention. 
This component refines the queries and renders the outputs.
The pixel decoder is made up of multiple upsampling CNN layers, and it is used to obtain per-pixel embeddings.
Given the representation $\mathcal{\hat{F}}$ and $N$ learnable queues $\{Q_0,Q_1...Q_{T}\}$ as input, the decoder outputs $N$ binary masks $\mathbf{O} = {o_1,\cdots,o_N} \in \{0,1\}^{N \times H \times W}$, along with the corresponding category. 
This is achieved through simple matrix multiplication between the outputs of the transformer decoder and the pixel decoder.
Following Mask2former~\cite{cheng2022masked}, one query is responsible for a class or instance for semantic and instance segmentation, respectively.
To optimize the mask prediction, we use the binary cross-entropy loss~\cite{cheng2021per} and dice loss~\cite{milletari2016v}. On the other hand, the cross-entropy loss is used for classification prediction.

\begin{figure}
    \centering
    \includegraphics[width=\linewidth]{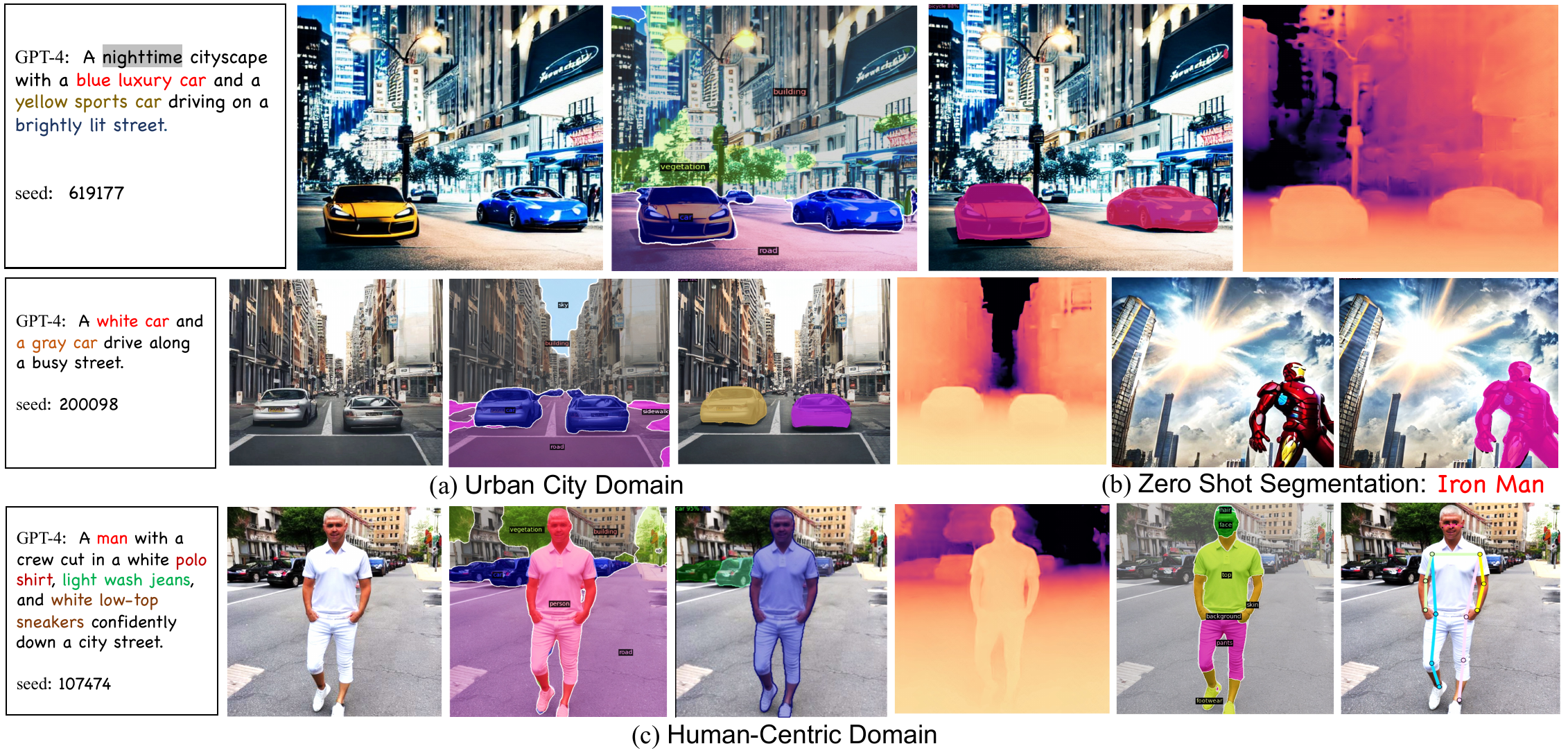}
    %%%\vspace{-4mm}
    \caption{
        \textbf{ Examples of generated data for \modelname.}
        Our method can produce semantic/instance segmentation, depth, and human pose annotation across various domains.
        }
    \label{fig:Examples}
    \vspace{-1mm}
\end{figure}

\textbf{Open-Vocabulary Segmentation.} Based on the architecture of generic segmentation, a semantics-related query concept is proposed to support the open-vocabulary segmentation task, as shown in Fig \ref{fig:decorder}-(b).
During the training stage, given the category name $\{C_0,C_1...C_{K}\}$ ~(\eg{}, \texttt{dog, cat}) on an image from the training set, we use the text encoder $\tau_\theta(\cdot)$, \eg{}, CLIP~\cite{radford2021learning} to encode them into the embedding space.
Then concatenating them with the queries to equip the semantics of class vocabulary to the query as follows:
\begin{align}
  \hat{Q}_i = \text{MLP}([Q_i,\tau_\theta(C_j)]),
\end{align}
where $Q_i$ and $C_j$ is the $i$-th query embedding and $j$-th class.
$\text{MLP}$ refers to a learned MLP, used to fuse the class embedding and learnable query embedding.
Thus \modelname can generate an open-vocabulary mask by incorporating a new class name, as illustrated in Fig. \ref{fig:Examples} (b).

\textbf{Depth and Pose Estimation.} For depth and pose estimation, the output format is predetermined, eliminating the need to differentiate between classes or instances.
In this context, the pixel decoder is only required to predict a fixed number of maps $\mathbf{O} \in \mathcal{R}^{M\times H \times W}$.
The value of $M$ is set to either $1$ or $17$ (corresponding to 17 human key points), depending on whether the task is depth or human pose estimation.
For human pose estimation, we use the mean squared error as the loss function, and the ground truth heatmaps are generated by applying a 2D Gaussian with a standard deviation of 1 pixel centered on each key point.
As for depth estimation, we update the loss function from the classic scale-invariant error~\cite{li2022depthformer,eigen2014depth}.

\begin{figure}
    \centering
    \includegraphics[width=\linewidth]{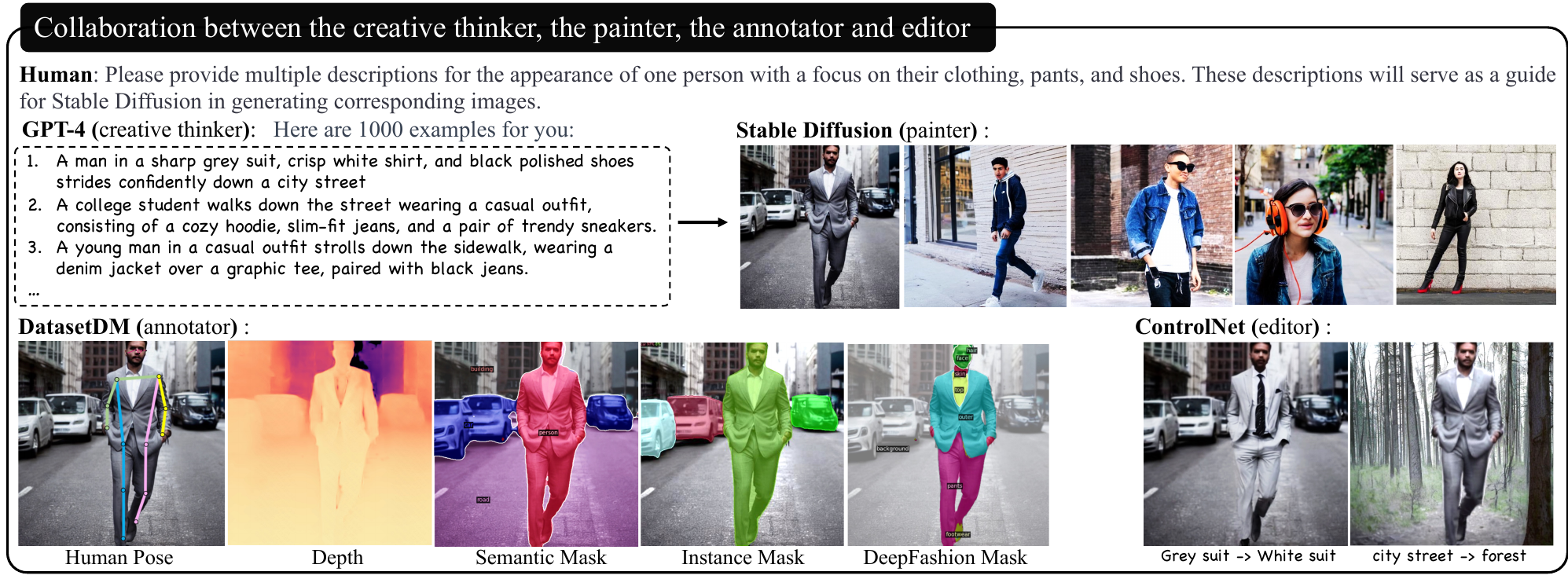}
    %%%\vspace{-4mm}
    \caption{
        \textbf{Collaboration between GPT-4 and Diffusion Model.
        }
        Large language models, \eg{}, GPT-4, can provide diverse and unrestricted prompts, enhancing the diversity of generated images.
        }
    \label{fig:Examples1}
    %%%\vspace{-1mm}
\end{figure}

\subsection{Prompting Text-Guided Data Generation}
\label{sec:Prompting}
In Fig. \ref{fig:pipeline} (b), we present the inference pipeline for text-guided data generation.
There are two main differences compared to the training phase: firstly, the prompts come from a large language model instead of a fixed template, and secondly, the denoising process is extended to $T$ steps to obtain the synthetic images.
Large language model, \ie{}, GPT-4, is adopted to enhance the diversity of generative data, while recent works~\cite{floridi2020gpt,hendy2023good,bubeck2023sparks} have proven their powerful understanding and adaptability for the real world.
As shown in Fig. \ref{fig:Examples1}, instead of the template-based prompts from humans, we guide GPT-4 to produce diverse, and infinite prompts.
For different downstream tasks and datasets, we give different guided prompts for GPT-4.
For example, as for the urban scene of Cityscapes~\cite{cordts2016cityscapes}, the simple guided prompt is like `\texttt{Please provide 100 language descriptions of urban driving scenes for the Cityscapes benchmark, containing a minimum of 15 words each. These descriptions will serve as a guide for Stable Diffusion in generating images.}'
In this approach, we collected $L$ text prompts, which average around 100 prompts for each dataset. 
For each inference, a random prompt is sampled from this set.
% , which is then used to seed the generation of each image.

% \begin{figure}
%     \centering
%     \includegraphics[width=\linewidth]{images/vis2.pdf}
%     \caption{
%         \textbf{ Examples of generated data %for
%         by 
%         \modelname.}
%         Our method can produce human pose and other annotations, such as masks across various different domains.
%         }
%     \label{fig:Examples2}
%     %%%\vspace{-1mm}
% \end{figure}

\section{Experiments}
\label{sec:experiments}
\subsection{Implementation details}
\textbf{Architecture and Training Details.}
Stable diffusion V1~\cite{rombach2022high} model pre-trained on the LAION5B~\cite{schuhmann2022laion} dataset is used as our text-to-image diffusion model.
The decoder architecture of Mask2Former~\cite{cheng2022masked} was selected as the base architecture for our P-Decoder. 
And we use $100$ queries for the segmentation task.
For all tasks, we train \modelname for around $50k$ iterations with images of size $512\times512$, which only need one Tesla V100 GPU, and lasted for approximately $20$ hours.
Optimizer~\cite{loshchilov2017decoupled} with a learning rate of $0.0001$ is used.
%

%

% \subsection{Implementation details}
% \label{sec:exp:impl}

\begin{table}[t]
    \centering
    \small 
    \setlength{\tabcolsep}{1mm}
    \centering

  \tablestyle{4pt}{1.2}
  \scriptsize
  % \footnotesize 
  \begin{tabular}[t]{p{0.3\columnwidth} p{0.18\columnwidth}| p{0.23\columnwidth} p{0.23\columnwidth} }
     \multicolumn{2}{c|}{Zero-Shot Segmentation}& \multicolumn{2}{c}{Long-Tail Segmentation} \\
     Seen Class & Unseen Class & Head Class & Tail Class \\
     \toprule
     aeroplane (0), bicycle (1), bird (2), boat (3), bottle (4), bus (5),  
     car (6), cat (7), chair (8), cow (9) , diningtable (10), dog (11), horse (12), motorbike (13), person (14)
     &
     potted plant (15), sheep (16), sofa (17), train (18), tvmonitor (19)
     &
    aeroplane (0), bicycle (1), bird (2), boat (3), bottle (4), bus (5),  
     car (6), cat (7), chair (8), cow (9) 
     &
     diningtable (10), dog (11), horse (12), motorbike (13), person (14),
     potted plant (15), sheep (16), sofa (17), train (18), tvmonitor (19)
\end{tabular}
    % %%%\vspace{-0.2cm}
    \caption{\textbf{Details for Zero-Shot and Long-tail Segmentation on VOC 2012~\cite{(voc)everingham2010pascal}.} }
    \label{zero}
    \vspace{-5mm}
\end{table}

\begin{table*}[t]
    \centering
    \small 
    \setlength{\tabcolsep}{1mm}
    \centering

  \tablestyle{4pt}{1.2}
  \scriptsize
  % \footnotesize 
  \begin{tabular}{l|cc|c|cc|c|cc|c|cc|c}
  & \multicolumn{3}{c|}{VOC~(Semantic Seg.)/\%} & \multicolumn{3}{c|}{COCO2017~(Instance Seg.)/\%}& \multicolumn{3}{c|}{NYU Depth V2~(Depth Est.)}& \multicolumn{3}{c}{COCO2017~(Pose Est.)/\%}\\
  method & \# real  & \# synth.  & mIoU & \# real  & \# synth.  & AP & \# real  & \# synth.  & REL~$\downarrow$ & \# real  & \# synth.  & AP  \\
  \shline
  Baseline & 100 & - & 65.2 & 400 & - & 14.4 & 50 & - & 0.31 & 800 & - & 42.4 \\
  \textbf{\modelname}
  & 100 & 40k  & 78.5 & 400 & 80k & 26.5 & 50 & 35k  & 0.21 & 800 & 80k  & 47.5\\

  \end{tabular}
    %%%\vspace{-0.2cm}
    \caption{\textbf{Downstream Tasks.} `real' and `synth.' denote real and synthetic images, respectively. The backbones of baselines for four tasks are `Swin-B', `Swin-B', `Swin-L', and `HR-W32', respectively.}
    % %%%\vspace{+2mm}
    \label{all}
    \vspace{-2mm}
\end{table*}

%%%%\vspace{-0.2cm}

\begin{table*}[t]
    \centering
    \small 
    \setlength{\tabcolsep}{1mm}
    \centering

  \tablestyle{4pt}{1.2}
  \scriptsize
  % \footnotesize 
  \begin{tabular}{l| l|cc | x{20}x{20}x{20}x{20}x{20}x{20}}
  method & backbone & \# real image & \# synthetic image & AP & AP$^\text{50}$ & AP$^\text{75}$ & AP$^\text{S}$ & AP$^\text{M}$ & AP$^\text{L}$  \\
  \shline
  Baseline & R50 & 400 & - & 4.4 & 9.5 & 3.5 & 1.1 & 3.3 & 12.1\\
  \textbf{\modelname} %(ours)
  & R50 & - & 80k~(R:400) &  12.2 & 24.3  & 10.9  & 1.6  & 11.3 & 30.9 \\
  \textbf{\modelname} %(ours)
  & R50 & 400 & 80k~(R:400) & 14.8 & 29.7 & 13.0 & 2.3 & 15.1  & 36.0 \\
  \hline
  Baseline & Swin-B & 400 & - & 11.3 & 23.0 & 9.6 & 3.2 & 10.1 & 27.1  \\
  \textbf{\modelname}  &  Swin-B & - & 80k~(R:400) & 17.6 & 34.1 & 15.8 & 3.4 & 17.8 & 39.5 \\
  \textbf{\modelname}  &  Swin-B & 400 & 80k~(R:400)& 23.3 & 43.0 & 22.2 & 7.7 & 26.1 & 48.7 \\
  \hline
  Baseline & Swin-B & 800 & - & 14.4 & 28.8 & 12.7 & 5.6 & 15.7 & 29.2 \\
  % \textbf{\modelname}  &  Swin-B & - & 80k~(R:800)& 19.3 & 35.2 & 18.7 & 3.5 & 20.4 & 43.9 \\
  \textbf{\modelname}  &  Swin-B & 800 & 80k~(R:800)& 26.5 & 46.9 & 25.8 & 7.7 & 29.8 & 53.3\\
  \end{tabular}
    %%%\vspace{-0.2cm}
    \caption{\textbf{Instance segmentation on COCO \texttt{val2017}.}  `R: ' denotes the real data used to train.}
    % %%%\vspace{+2mm}
    \label{COCO_Ins}
    \vspace{-2mm}
\end{table*}
\textbf{Downstream Tasks Evaluation.}
To comprehensively evaluate the generative image of \modelname, we conduct seven groups of experiments for the supported six downstream tasks.
\textit{Semantic Segmentation.}
Pascal-VOC 2012~\cite{(voc)everingham2010pascal} (20 classes) and Cityscapes~\cite{(cityscape)cordts2016cityscapes}~(19 classes), as two classical benchmark are used to evaluate.
We synthesized $2k$ images for each class in both datasets, resulting in a total of $40k$ and $38k$ synthetic images for Pascal-VOC 2012 and Cityscapes, respectively.
The synthetic data is subsequently utilized to train Mask2former~\cite{cheng2022masked} and compared to its real data counterpart on a limited dataset setting~(around $100$ images).
\textit{Instance Segmentation.} For the COCO2017~\cite{lin2014microsoft} benchmark, we synthesized $1k$ images for each class, resulting in a total of $80k$ synthetic images. 
Similarly, Mask2former~\cite{cheng2022masked}, as the baseline, is used to evaluate the synthetic data.
We evaluate only the class-agnostic performance, where all the $80$ classes are assigned the same class ID.
\textit{Depth Estimation.} We synthesized a total of $80k$ synthetic images for NYU Depth V2~\cite{silberman2012indoor}.
And using Depthformer~\cite{li2022depthformer}\footnote{\tt https://github.com/zhyever/Monocular-Depth-Estimation-Toolbox} to evaluate our synthetic data.
\textit{Pose Estimation.} We generated a set of $30k$ synthetic images for COCO2017 Pose dataset~\cite{lin2014microsoft} and employed HRNet~\cite{sun2019deep} as the baseline model to assess the effectiveness of our approach.
\textit{Zero-Shot Semantic Segmentation.}
Following Li \textit{et al.}~\cite{li2023guiding}, Pascal-VOC 2012~\cite{(voc)everingham2010pascal} (20 classes) is used to evaluate.
We train \modelname with only 15 seen categories, where each category including 30 real images, and synthesized a total of $40k$ synthetic images for 20 categories.
\textit{Long-tail Semantic Segmentation.}
The categories of VOC 2012 are divided into head (20 images each class) and tail
classes (2 images each class).
Then we train \modelname with these data and generate synthetic data.
\textit{Human Semantic Segmentation.} We synthesized a total of $20k$ synthetic images for DeepFashion-MM~\cite{jiang2022text2human}~(24 classes).
Mask2former~\cite{cheng2022masked} is used to evaluate the synthetic data.
We split DeepFashion-MM
into a set of 100 training images, and 12,500 testing images.
Further details, such as prompts, can be found in the supplementary material.

\textbf{Class Split for Zero-Shot and Long-Tail Segmentation.}
Table   \ref{zero} provides a comprehensive overview of the class distribution for both zero-shot and long-tail scenarios. The division for zero-shot classes is consistent with previous studies~\cite{bucher2019zero,wu2023diffumask,li2023guiding}. 
The configuration for long-tail data distribution is firstly established in this paper.

\begin{table}[t]
    \centering
    \small 
    \setlength{\tabcolsep}{1mm}
    \centering

  \tablestyle{4pt}{1.2}\scriptsize\begin{tabular}{l| l|cc | x{20}x{20}x{20}x{20}x{20}| x{20}}
   &  &  &  & \multicolumn{5}{c|}{Sampled Classes for Comparison/\%}& \\
  method & backbone & \# real image & \# synthetic image & bird & cat & bus & car & dog & mIoU  \\
  \shline
  Baseline & R50 & 100 & - & 54.8 & 53.3 & 69.3 & 66.8 & 24.2 & 43.4 \\
  \textbf{\modelname} (ours) & R50 & - & 40k~(R:100) & 84.7  & 74.4 &  86.0 & 79.2 & 63.7   & 60.3\\
  \textbf{\modelname} (ours) & R50 & 100 & 40k~(R:100) &  81.7 & 82.3 & 87.7 & 77.9 & 69.3 & 66.1 \\
  \hline
  Baseline & Swin-B & 100 & - & 54.4 & 68.3 & 86.5& 71.8 & 49.1 & 65.2 \\
  \textbf{\modelname} (ours) &  Swin-B & - & 40k~(R:100) & 93.4& 94.5 & 93.8 & 78.8 & 79.6 & 73.7\\
  \textbf{\modelname} (ours) &  Swin-B & 100 & 100~(R:100) & 83.9& 71.0 & 82.9 & 78.0 & 39.5 & 67.9\\
  \textbf{\modelname} (ours) &  Swin-B & 100 & 400~(R:100) & 86.9& 92.0 & 90.8 & 82.6 & 86.7 & 76.1\\
  \textbf{\modelname} (ours) &  Swin-B & 100 & 40k~(R:100)& 86.7 & 93.8 & 92.3& 88.3& 87.1 & 78.5 \\
  \hline
  Baseline & Swin-B & 10.6k  (full) & - & 93.7 & 96.5 & 90.6 & 88.6 & 95.7 & 84.3 \\
  \textbf{\modelname} (ours) &  Swin-B & 10.6k (full) & 40k~(R:100)& 93.9 & 97.6 & 91.9& 89.4& 96.1 & 85.4 \\
  \end{tabular}

  % \centering

  % \tablestyle{4pt}{1.2}\scriptsize\begin{tabular}{l| ccc|cc | x{20}}
  %  % &  &  &  & \multicolumn{5}{c|}{Sampled Classes for Comparison/\%}& \\
  % method & baseline & data aug. & backbone & \# real image & \# synthetic image & mIoU  \\
  % \shline
  % - & Mask2former & crop\&color R50 & 100 & -  & 43.4 \\
  % \modelname & Mask2former & crop\&color & R50 & - & 40k~(R:100)  & 60.3\\
  % \modelname & Mask2former & crop\&color & R50 & 100 & 40k~(R:100)  & 66.1 \\
  % \hline
  % - & Mask2former & crop\&color & Swin-B & 100 & - & 65.2 \\
  % \modelname & Mask2former & crop\&color &  Swin-B & - & 40k~(R:100) & 73.7\\
  % \modelname & Mask2former & crop\&color &  Swin-B & 100 & 40k~(R:100)& 78.5 \\
  % \end{tabular}
    % %%%\vspace{-0.2cm}
    \caption{\textbf{Semantic segmentation on VOC  2012.}  `R: ' refers to the number of real data used to train.}
    % %%%\vspace{+2mm}
    \label{VOC_sem}
    \vspace{-2mm}
\end{table}

\begin{table}[t]
    \centering
    \small 
    \setlength{\tabcolsep}{1mm}
    \centering

  \tablestyle{4pt}{1.2}\scriptsize\begin{tabular}{l| l|cc | x{20}| x{20}| x{20} x{20} x{20}| x{20}}
  
  & & & &\multicolumn{2}{c|}{8 Classes/\%}&\multicolumn{4}{c}{19 Classes/\%}\\
  method & backbone & \# real image & \# synthetic image & vehicle & mIoU & car & bus & bicycle & mIoU  \\
  \shline
  Baseline & R50 & 9 & 100k+~(R:16) & 84.3 &  71.5& 82.8 & 22.3 & 42.4 & 36.8 \\
  HandsOff~\cite{xu2022handsoff} & R101 & 16 & 100k+~(R:16) & -& 55.1 &- & - & -& - \\
  HandsOff~\cite{xu2022handsoff} & R101 & 50 & 100k+~(R:16) & -& 60.4 & -&  -&- &-  \\
  \textbf{\modelname} (ours) & R50 & - & 38k~(R:9) &  86.9 & 69.5& 83.3 & 8.3 & 53.5 & 34.2 \\
  \textbf{\modelname} (ours) & R50 & 9 & 38k~(R:9) & 88.6 &76.7  & 85.6 & 28.9 & 56.5 &42.1\\
  \textbf{\modelname} (ours) & R101 & 9 & 38k~(R:9) & 88.9 & 77.5 & 85.9 & 27.9 & 60.4 & 43.7\\
  \hline
  Baseline & Swin-B & 9 & - &  84.1 &  74.5 & 83.3 & 27.7 & 42.0 & 41.1\\
  \textbf{\modelname} (ours) &  Swin-B & - & 38k~(R:9) & 85.7 & 73.3 & 84.3& 20.3& 29.1& 37.3\\
  \textbf{\modelname} (ours) &  Swin-B & 9 & 38k~(R:9) & 89.4 & 80.0 & 87.2& 30.2& 66.5 & 47.4 \\
  \end{tabular}
    % %%%\vspace{-0.2cm}
    \caption{\textbf{Semantic segmentation on Cityscapes for two different split settings: $8$ and $19$ categories.}  `vehicle', `car', `bus', and `bicycle' are sampled classes for presentation.}
    % %%%\vspace{+2mm}
    \label{cityscapes_sem}
    % %%%\vspace{-2mm}
\end{table}

\begin{table}[t]
    \centering
    \small 
    \setlength{\tabcolsep}{1mm}
    % \begin{subtable}{0.49\linewidth}
%   \centering
%   \tablestyle{3pt}{1.2}
%   \footnotesize
%   \begin{tabular}{cc|ccc}
%   Dynamic Query & Query G-KD & 1-15 & 16-20 & \textit{all} \\
%   \shline
%   &  & 69.1 & 30.5 & 59.4 \\
%   \checkmark &  &  &  &   \\
%   \checkmark & \checkmark & 85.2 \dt{+16.1} & 37.9 \dt{+7.4} & 73.4 \dt{10.4}\\
% %   \multicolumn{5}{c}{~}\\
%   \end{tabular}
%   \caption{\textbf{Dynamic Query and Query Guided KD}. Experiments are conducted on \textit{15-1} setting for semantic segmentation~(mAP) task on VOC \texttt{val}.
%   }
%   \label{ablation_DQ1}
%   \end{subtable}\hspace{2mm}
  
  \begin{subtable}{0.23\linewidth}
  \centering
  \tablestyle{2.3pt}{1.2}
  \scriptsize
  \begin{tabular}{ r |cc|c}
   step & car & dog & \textit{mIoU} \\
  \shline
   1 &  88.3 & 87.1 & 78.5 \\
   100 &  88.1 & 87.2 & 78.5  \\
   200 &  88.0 & 86.6 & 78.3 \\
   500 &  87.2 & 84.9 & 76.8 \\
   800 &  86.3 & 83.4 & 76.1
  \end{tabular}
  \caption{\textbf{Visual Features $\mathcal{F}$}.
  }
  \label{ablation_timesetp}
  \end{subtable}\hspace{0.3mm}
  \begin{subtable}{0.21\linewidth}
  \centering
  \tablestyle{2.3pt}{1.2}
%   \scriptsize
  \scriptsize
  \begin{tabular}{c|cc|c}
  step &  car & dog & \textit{mIoU} \\
  \shline
   - & 88.1 & 87.0 & 77.6\\
   1 &  88.3 & 87.1 & 78.5 \\
   200 & 87.7  & 87.0 & 78.0\\
    500 &  87.2 & 86.3 & 77.5\\
   800 &  87.1 & 86.1 & 77.1
  \end{tabular}
  \caption{\textbf{Cross Attention $\mathcal{\hat{A}}$.} 
  }
  \label{tab:ablation:attention}
  \end{subtable}
  \hspace{0.3mm}
  \begin{subtable}{0.22\linewidth}
  \centering
  \tablestyle{2.3pt}{1.2}
  \scriptsize
  \begin{tabular}{c|cc}
  % & \multicolumn{2}{c}{\textit{mIoU}/\%}\\
  \# train im. & syn. & joint \\
  \shline
  60  & 71.4  &  77.1   \\
  100  & 73.7  & 78.5    \\
  200 & 74.4 & 79.4 \\
  400 & 76.4 &  80.4  \\ 
  1,000 & 78.4 & 81.1     
  
%   \multicolumn{4}{c}{~}\\
  \end{tabular}
  \caption{\textbf{Size of Train Set.}
  }
  \label{ablation_NL}
  \end{subtable}
  \hspace{0.3mm}
  \begin{subtable}{0.25\linewidth}
  \centering
  \tablestyle{2.3pt}{1.2}
  \scriptsize
  \begin{tabular}{l|cc|c}
  prompt~(\# num.) & car & dog  &\textit{mIoU} \\
  \shline
  Human~(100)   & 84.9 & 84.5 & 76.6 \\
  GPT-4~(100)  & 85.2 & 86.1 &77.1 \\
  GPT-4~(200) & 88.0 & 86.2 & 77.3 \\
  GPT-4~(500) & 88.1 & 87.1 & 78.5\\
  GPT-4~(1k) & 88.3 & 87.1 & 78.5
  \end{tabular}
  \caption{\textbf{Prompt Candidates.}
  }
  \label{ablation_param}
  \end{subtable}

    %%%\vspace{-0.1cm}
    \caption{\textbf{\modelname Ablation on Pascal-VOC 2012 for semantic segmentation.}  Swin-B is used as the backbone. 100 real images are used for (a), (b), and (d). `Syn.' and `Joint' denote training with only synthetic data and joint training with real data, respectively.}
    % %%%\vspace{+2mm}
    \label{ablation_study}
    \vspace{-2mm}
\end{table}

\subsection{Main Results}
Table   \ref{all} provides a basic comparison of the selected four downstream tasks. 
More additional experiments can be found in Table   \ref{COCO_Ins} and Table   \ref{VOC_sem}, as well as in the Supplementary Material (\ie{}, Pose Estimation, Depth Estimation, Zero-Shot Segmentation, Fashion Segmentation, and others).

\textbf{Semantic Segmentation.} Table   \ref{VOC_sem} displays the performance on VOC 2012.
Using only 100 real images (5 images per class), training with purely synthetic data from \modelname achieves a $73.7\%$ mIoU, an $8.5\%$ improvement compared to using real data alone.
Moreover, when jointly training with the 100 real images, further improvement is observed, resulting in a mIoU of $78.5\%$.
Table~\ref{cityscapes_sem} displays the performance on Cityscapes~\cite{(cityscape)cordts2016cityscapes}.
To compare with previous methods~\cite{xu2022handsoff}, we also conducted experiments with a 9-classes division. 
%
% Furthermore, we also provided experiments with a 19-class division, which presents a more challenging setting.
%
\modelname demonstrates consistent advancements over the baseline or prior SOTA, achieving up to a $10\%$ improvement in mIoU under each experimental setup.

\textbf{Instance Segmentation.} Table   \ref{COCO_Ins} presents three distinct training settings, encompassing variations in the backbone and the number of training images. 
Regardless of the chosen setting, \modelname consistently achieves an improvement of approximately $10\%$. Employing 800 training images (10 images per class) and the Swin-B backbone, \modelname yields a $12.1\%$ increase in Average Precision (AP), resulting in a final AP of $26.5\%$.

\textbf{Depth Estimation}. Table   \ref{all} presents a concise comparison between synthetic and real data on the NYU Depth V2 dataset~\cite{silberman2012indoor}. Detailed information (\eg{}, backbone, other metrics) can be found in the supplementary material.
When trained with 50 images, \modelname can achieve a $10\%$ improvement compared to training solely with real images.

\textbf{Human Pose Estimation.} For the human pose estimation task on the COCO2017 dataset, \modelname demonstrates significant improvements compared to the baseline trained on 800 real images, achieving a $5.1\%$ increase, as illustrated in Table   \ref{all}.
Similar to depth estimation, additional information can be found in the supplementary material.

\textbf{Zero Shot and Long-tail Segmentation.} 
Table   \ref{fig:zeroshot} displays the results of experiments related to zero-shot and long-tail segmentation. 
Our model, \modelname, notably alleviates the effects of long-tail distribution by synthesizing a substantial amount of data for rare classes, leading to an improvement of up to $20\%$ in mIoU.
Details for both tasks can be found in the supplementary material.

%%%%\vspace{-5mm}
\subsection{Ablation studies}

%%%%\vspace{-5mm}
\begin{minipage}{0.6\textwidth}
\begin{table}[H] % Add [H] to keep the table in place
\scriptsize
\begin{tabular}{c|cc|c|cc|c}
% \hline
 & \multicolumn{3}{c|}{Zero-Shot Setting}& \multicolumn{3}{c}{Long-tail Setting} \\
Method & seen & unseen & harm. & head & tail & mIoU\\
\hline
Baseline(no Syn.) & 61.3 & 10.7 & 18.3 & 61.2 & 44.1 &52.6  \\
Li \textit{et al.}~\cite{li2023guiding} & 62.8 & 50.0 & 55.7 & -& -& - \\
DiffuMask~\cite{wu2023diffumask} & 71.4& 65.0 &68.1 & -& -& - \\
\modelname & 78.8 & 60.5& 68.4 & 73.1& 66.4& 70.0 \\

\end{tabular}
\vspace{.2cm}
\caption{\textbf{Zero Shot and Long-tail Segmentation on VOC 2012.} Zero Shot: following priors~\cite{li2023guiding,wu2023diffumask},
we train \modelname with only 15 seen categories, and tested for 20 categories.
Long-tail Setting: the 20 categories are divided into head~(10 classes, 20 images each class) and tail classes~(10 classes, 2 images each class).} % 
\label{fig:zeroshot}
\end{table}

\end{minipage}
\begin{minipage}{0.4\textwidth}
\begin{figure}[H] % Add [H] to keep the figure in place
\centering
\begin{tikzpicture}
\begin{axis}[
  xlabel={Word Number of Prompt},
  ylabel={mIoU},
  xmin=5, xmax=30,
  ymin=70, ymax=95,
  xtick={5, 10,15,20,25,30},
  ytick={70,75,80,85,90,95},
  legend pos=south east,
  ymajorgrids=true,
  grid style=dashed,
  xlabel near ticks,
  ylabel near ticks,
  width=0.99\textwidth,
  height=0.7\textwidth,
  xlabel style={font=\footnotesize}, % Adjust x-axis label font size
  ylabel style={font=\footnotesize}, % Adjust y-axis label font size
  xticklabel style={font=\tiny}, % Adjust x-axis tick label font size
  yticklabel style={font=\tiny}, % Adjust y-axis tick label font size
  legend style={font=\tiny}, % Adjust legend (annotation) font size
]

\addplot[
  color=blue,
  mark=square,
]
coordinates {
  (5,78.5)(10,86.7)(15,85.4)(20,86.4)(25,87.4)(30,86.6)
};
\addlegendentry{Bird}

\addplot[
  color=red,
  mark=triangle,
]
coordinates {
  (5,89.3)(10,93.9)(15,93.5)(20,93.3)(25,93.6)(30,93.2)
};
\addlegendentry{Cat}

\addplot[
  color=green,
  mark=asterisk,
]
coordinates {
  (5,81.9)(10,86.9)(15,87.1)(20,87.0)(25,87.3)(30,87.1)
};
\addlegendentry{Person}

\end{axis}
\end{tikzpicture}
%%%%\vspace{-2mm}
\caption{\textbf{Ablation of Prompt Length.}} % Add the caption

\label{fig:line_plot}
\end{figure}
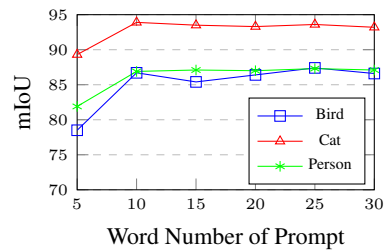
\end{minipage}

\textbf{Diffusion Time Steps.} Table   \ref{ablation_timesetp} depicts the influence of visual features derived from various diffusion time steps, and the maximum sampling step is 1000.
We observe that a large step results in adverse outcomes, whereas the performance with a smaller step tends to remain relatively stable.
%
% Moreover, w
% When the step size is less than 500, the performance remains approximately consistent.
%

\textbf{Cross-Attention Fusion.} 
Fig. \ref{tab:ablation:attention} demonstrates that the cross attention maps, $\mathcal{F}$ in step 1 can yield a modest improvement, roughly $1\%$.
Interestingly, it appears that as the step size increases, the benefit becomes less pronounced.
Indeed, when the step size surpasses 500, it may even result in detrimental effects.
From this perspective, the utility of the cross-attention map is limited.

\textbf{Training Set Size.} Additional training data for \modelname can further improve synthetic data, as shown in Table   \ref{ablation_NL}. 
The increase of the training data from 60 to 400 images precipitates the most conspicuous improvement, subsequently reaching a saturation point.
With 1k training images, the performance escalates to an impressive $81\%$, demonstrating competitive prowess for the application. 
Notably, 1k training images representing roughly $10\%$ of the original data, is still relatively diminutive.

\textbf{Prompt from Language Model}. \textit{Candidate Number.} We also investigated the impact of the number of prompt candidates, as depicted in Table   \ref{ablation_param}.
%
% We directed GPT-4 to generate several groups of prompts, each containing varying quantities of prompt languages.
%
With the current configurations, an increase in the number of prompts can potentially lead to a performance improvement of $2\%$.
\textit{Word Number of Each Prompt.}
%
% The length and content of prompt language can be modulated by guiding GPT-4~(refer to the supplementary material).
%
We simply study the effect of the length of prompt in Fig. \ref{fig:line_plot}.
An increase in text length from 5 to 10 yields approximately $4\%$ enhancement. 
When the text length surpasses $10$, the performance appears to plateau. 
We argue that the upper limit is due to the current capacity of generative models. 
%
% Employing more powerful generative models might yield additional improvements.

\section{Conclusion}
In this study, we investigate using a perception decoder to parse the latent space of an advanced diffusion model, extending the text-to-image task to a new paradigm: text-guided data generation.
Training the decoder requires less than $1\%$ of existing labeled images, enabling infinite annotated data generation.
Experimental results show that the existing perception models trained on synthetic data generated by DatasetDM exhibit exceptional performance across six datasets and five distinct downstream tasks. 
Specifically, the synthetic data yields significant improvements of $13.3\%$ mIoU for semantic segmentation on VOC 2012 and $12.1\%$ AP for instance segmentation on COCO 2017.
Furthermore, text-guided data generation offers additional advantages, such as a more robust solution for domain generalization and enhanced image editing capabilities. 
We hope that this research contributes new insights and fosters the development of synthetic perception data.

\section*{Acknowledgements}
% This work was in part supported by the National Key R\&D Program of China (No.\ 2022ZD0118700).

W. Wu, C. Shen's participation was 
 supported by the National Key R\&D Program of China (No.\  2022ZD0118700). 
W. Wu, H. Zhou's participation was  supported by the National Key Research and Development Program of China (No.\ 2022YFC3602601), and the Key Research and Development Program of Zhejiang Province of China (No.\ 2021C02037).
%
% M. Shou's participation was  supported by the National Research Foundation, Singapore under its NRFF Award NRF-NRFF13-2021-0008, and his Start-Up Grant from National University of Singapore.
%
M. Shou's participation was  supported by the National Research Foundation, Singapore under its NRFF Award NRF-NRFF13-2021-0008, and the Ministry of Education, Singapore, under the Academic Research Fund Tier 1 (FY2022).

\appendix

\section{Implementation details}
\subsection{Dataset Details}
\begin{itemize}
     \itemsep 0.0cm 
    \item Pascal-VOC 2012~\cite{(voc)everingham2010pascal} (20 classes) is a popular dataset for semantic segmentation in computer vision. 
    It contains thousands of annotated images featuring 20 different object classes, such as animals, vehicles, and furniture. 

    \item Cityscapes~\cite{cordts2016cityscapes}~(19 classes) is a benchmark dataset for semantic urban scene, primarily focusing on semantic segmentation tasks in computer vision. 
    It contains high-quality pixel-level annotations of 5,000 images from 50 different cities, captured at various times of the day and under diverse weather conditions.
    There are 2,975 images for training and 500 images for validation.

    \item COCO 2017 (Common Objects in Context)~\cite{lin2014microsoft} is a widely-used benchmark dataset for computer vision tasks, such as object detection, segmentation, and human pose estimation.
    It contains over 200,000 labeled images with 1.5 million object instances belonging to 80 object categories

    \item NYU Depth V2~\cite{silberman2012indoor} is designed for indoor scene understanding tasks in computer vision, specifically for depth estimation task. 
    The NYU Depth V2 dataset contains 1,449 labeled images and 407,024 unlabeled frames, captured from 464 diverse indoor scenes.

    \item DeepFashion-MM~\cite{jiang2022text2human}~(24 classes) is a benchmark dataset designed for the task of clothing synthesis in the field of computer vision. It consists of 24 different clothing classes.

\end{itemize}

\begin{table}[t]
    \centering
    \small 
    \setlength{\tabcolsep}{1mm}
    \centering

  \tablestyle{4pt}{1.2}\scriptsize\begin{tabular}{l| c| c|c |c}
  Task & Dataset & Full Real Data & Used for \modelname & \# synthetic image   \\
  \shline
  Instance Segmentation & COCO 2017~\cite{lin2014microsoft} & 118.3k & 400 (0.3\%) & 80k    \\
  Semantic Segmentation & VOC 2012~\cite{(voc)everingham2010pascal} & 10.6k & 100 (0.87\%) & 40k    \\
  Semantic Segmentation & Cityscapes~\cite{cordts2016cityscapes} & 2.9k & 9 (0.3\%) & 38k    \\
  Semantic Segmentation & DeepFashion-MM~\cite{jiang2022text2human} & 12.7k & 120 (0.9\%) & 38k    \\
  Zero-Shot Segmentation & VOC 2012~\cite{(voc)everingham2010pascal} & 10.6k & 450 (3.9\%) & 40k    \\
  % Depth & KITTI & 23.2k & 50 (0.2\%) & 200k    \\
  Depth & NYU Depth V2~\cite{silberman2012indoor} & 24.2k & 50 (0.2\%) & 35k    \\
  Human Pose &  COCO 2017-Pose~\cite{lin2014microsoft} & 118.3k & 800 (0.6\%) & 80k    \\
  \end{tabular}
\vspace{0.2cm}
    \caption{\textbf{Comparison of data size.}  With less than 1\% manually labeled images, \modelname can enable the generation of an infinitely large annotated dataset.}
    % %%%\vspace{+2mm}
    \label{data}
    \vspace{-2mm}
\end{table}

\begin{table*}[t]
    \centering
    \small 
    \setlength{\tabcolsep}{1mm}
    \centering

  \tablestyle{4pt}{1.2}
  \scriptsize
  % \footnotesize
  \begin{tabular}{l| l|l|cc | x{20}x{20}x{20}x{20}x{20}x{20}}
  method & backbone & input size & \# real im. & \# synthetic im. & AP & AP$^\text{50}$ & AP$^\text{75}$ & AP$^\text{M}$ & AP$^\text{L}$ & AR  \\
  \shline
  Baseline & R50 & 256 $\times$ 192 & 800 & - & 31.3 & 62.0 & 27.7 & 30.7 & 32.0 & 36.2 \\
  \textbf{\modelname}  & R50 & 256 $\times$ 192 &- & 80k~(R:800) & 11.4 & 28.2 & 7.8 & 6.9 & 17.7 & 14.3 \\
  \textbf{\modelname}  & R50 & 256 $\times$ 192 & 800 & 80k~(R:800) & 36.4& 66.7& 35.0 & 33.0 & 40.8 & 40.1 \\
  \hline
  Baseline & HR-W32 & 256 $\times$ 192& 800 & - & 42.4 & 73.3 & 42.1 & 39.5 & 47.0 & 46.7  \\
  \textbf{\modelname} & HR-W32 &256 $\times$ 192 & - & 80k~(R:800) &  13.4 & 30.9 & 9.9 & 8.0 & 21.7& 17.7  \\
  \textbf{\modelname} & HR-W32 &256 $\times$ 192 & 800 & 80k~(R:800) & 47.5 & 75.6&  49.3& 44.2 & 52.6 & 51.2 \\
  \hline
  Baseline & HR-W32 & 384 $\times$  288 & 800 & - & 43.4 & 72.2 & 44.7 & 40.5 & 47.9 & 47.5\\
  \textbf{\modelname} & HR-W32 &384 $\times$  288  & 800 & 80k~(R:800) & 48.9 & 76.7 & 51.4 & 44.6 & 55.0 &52.4 \\
  \end{tabular}
    \vspace{-0.2cm}
    \caption{\textbf{Human Pose Estimation on COCO \texttt{val2017}.}  `R: ' refers to the number of real data used for training \modelname.}
    % %%%\vspace{+2mm}
    \label{COCO_Pos}
    \vspace{-2mm}
\end{table*}

\begin{table}[t]
    \centering
    \small 
    \setlength{\tabcolsep}{1mm}
    \centering

  \tablestyle{4pt}{1.2}
  \scriptsize
  % \footnotesize 
  \begin{tabular}[t]{l l}
     Collapsed label (8) & Cityscapes (Fine annotations) original labels  \\
     \toprule
     Void & \begin{tabular}{@{}l@{}}Unlabeled (0), ego vehicle (1), rectification border (2), out of ROI (3), static (4), dynamic (5),  
     ground (6), \\ sidewalk (8), parking (9), rail track (10)\end{tabular}\\

     Road & Road (7) \\
     Construction & Building (11), wall (12), fence (13), guard rail (14), bridge (15), tunnel (16)\\
     Object & pole (17), polegroup (18), traffic light (19), traffic sign (20)\\
     Nature & Vegetation (21), terrain (22)\\
     Sky & Sky (23)\\
     Human & Person (24), rider (25)\\
     Vehicle & \begin{tabular}{@{}l@{}}UCar (26), truck (27), bus (28), caravan (29), trailer (30), train (31), motorcycle (32),
     bicycle (33), license plate (-1)\end{tabular}\\ \\
\end{tabular}
    %%%\vspace{-0.2cm}
    \caption{\textbf{Details for $8$ and $19$ categories on Cityscapes~\cite{cordts2016cityscapes}}.}
    % %%%\vspace{+2mm}
    \label{cityscapes_class}
    \vspace{-3mm}
\end{table}

\begin{table}[t]
    \centering
    \small 
    \setlength{\tabcolsep}{1mm}
      \centering

  \tablestyle{4pt}{1.2}\scriptsize\begin{tabular}{lc| c|cc | x{20}x{20}}
   % &  &  &  & \multicolumn{5}{c|}{Sampled Classes for Comparison/\%}& \\
  data aug. & baseline & backbone & \# real image & \# synthetic image & mIoU & Improv. \\
  \shline
  crop & Mask2former~\cite{cheng2022masked} & R50 & 100 & -  & 41.5 & - \\
  flip, crop, color & Mask2former~\cite{cheng2022masked} & R50 & 100 & -  & 43.4 & +1.9\\
  crop, \modelname & Mask2former~\cite{cheng2022masked} & R50 & 100 & 40k~(R:100)  & 65.2 & +23.7\\
  flip, crop, color, \modelname & Mask2former~\cite{cheng2022masked} & R50 & 100 & 40k~(R:100)  & 66.1 & +24.6\\
  \hline
  crop & Mask2former~\cite{cheng2022masked} & Swin-B & 100 & -  & 64.1& - \\
  flip, crop, color & Mask2former~\cite{cheng2022masked} & Swin-B & 100 & - & 65.2 &+1.1 \\
  crop, \modelname & Mask2former~\cite{cheng2022masked}  & Swin-B & 100 & 40k~(R:100)& 77.8&+13.7\\
  flip, crop, color, \modelname & Mask2former~\cite{cheng2022masked}  & Swin-B & 100 & 40k~(R:100)& 78.5 & +14.4\\
  \hline
  crop & DeepLabV3+~\cite{chen2018encoder} & Mobilenet & 100 & -  & 39.1 & - \\
  crop, \modelname & DeepLabV3+~\cite{chen2018encoder} & Mobilenet & 100 & 40k~(R:100)& 45.3 & +6.2   \\
  flip, crop, color & DeepLabV3+~\cite{chen2018encoder} & Mobilenet & 100 & - &  40.5 & +1.4\\
  flip, crop, color, \modelname & DeepLabV3+~\cite{chen2018encoder} & Mobilenet & 100 & 40k~(R:100)& 46.1 &+7.0 \\
  \hline
  crop & DeepLabV3+~\cite{chen2018encoder} & R50 & 100 & -  & 45.1 & -   \\
  crop, \modelname & DeepLabV3+~\cite{chen2018encoder} & R50 & 100 & 40k~(R:100)& 55.3& +10.2  \\
  flip, crop, color & DeepLabV3+~\cite{chen2018encoder} & R50 & 100 & - & 46.3 & +1.2\\
  flip, crop, color, \modelname & DeepLabV3+~\cite{chen2018encoder} & R50 & 100 & 40k~(R:100)& 56.9 &+11.8 
  \end{tabular}

  %\centering

  % \tablestyle{4pt}{1.2}\scriptsize\begin{tabular}{l| l|cc | x{20}x{20}x{20}x{20}x{20}| x{20}}
  %  &  &  &  & \multicolumn{5}{c|}{Sampled Classes for Comparison/\%}& \\
  % method & backbone & \# real image & \# synthetic image & bird & cat & bus & car & dog & mIoU  \\
  % \shline
  % Baseline & R50 & 100 & - & 54.8 & 53.3 & 69.3 & 66.8 & 24.2 & 43.4 \\
  % \textbf{\modelname} (ours) & R50 & - & 40k~(R:100) & 84.7  & 74.4 &  86.0 & 79.2 & 63.7   & 60.3\\
  % \textbf{\modelname} (ours) & R50 & 100 & 40k~(R:100) &  81.7 & 82.3 & 87.7 & 77.9 & 69.3 & 66.1 \\
  % \hline
  % Baseline & Swin-B & 100 & - & 54.4 & 68.3 & 86.5& 71.8 & 49.1 & 65.2 \\
  % \textbf{\modelname} (ours) &  Swin-B & - & 40k~(R:100) & 93.4& 94.5 & 93.8 & 78.8 & 79.6 & 73.7\\
  % \textbf{\modelname} (ours) &  Swin-B & 100 & 40k~(R:100)& 86.7 & 93.8 & 92.3& 88.3& 87.1 & 78.5 \\
  % \end{tabular}
    % %%%\vspace{-0.2cm}
    \caption{\textbf{Comparison with Data Augmentation.}  `R: ' refers to the number of real data used to train. `crop', `flip', and `color' refer to the `random crop', `random horizontal flip', and `color augmentation', respectively.}
    % %%%\vspace{+2mm}
    \label{ablation_aug_baseline}
    \vspace{-3mm}
\end{table}

\begin{table}[t]
    \centering
    \small 
    \setlength{\tabcolsep}{1mm}
    \centering

  \tablestyle{4pt}{1.2}\scriptsize\begin{tabular}{l| l|cc | x{20}x{20}x{20}x{20}x{20}| x{20}}
   &  &  &  & \multicolumn{5}{c|}{Sampled Classes for Comparison/\%}& \\
  method & backbone & \# real image & \# synthetic image & outer & dress & headwear & belt & socks & mIoU  \\
  \shline
  Baseline & R50 & 100 & - &58.2 & 65.2 & 19.2 & 24.3 & 0 & 31.2 \\
  \textbf{\modelname} (ours) & R50 & - & 40k~(R:100) & 53.1 & 57.2 & 0.4 & 0.4 & 0 & 28.9\\
  \textbf{\modelname} (ours) & R50 & 100 & 40k~(R:100) & 53.1 & 59.3 & 34.3 & 59.1 & 3.2 & 36.7\\
  \hline
  Baseline & Swin-B & 100 & - & 58.1 & 56.1 & 64.3 & 33.4 & 7.2 & 40.1\\
  % \textbf{\modelname} (ours) &  Swin-B & - & 40k~(R:100) &  &  &  &  & \\
  \textbf{\modelname} (ours) &  Swin-B & 100 & 40k~(R:100) & 70.0 & 70.8 & 72.0 & 32.8 & 5.9 & 45.1\\
  \end{tabular}
    % %%%\vspace{-0.2cm}
    \caption{\textbf{Semantic segmentation on DeepFashion-MM~\cite{jiang2022text2human}.}  `R: ' refers to the number of real data used to train.}
    % %%%\vspace{+2mm}
    \label{deepfashion_sem}
    \vspace{-2mm}
\end{table}

\begin{table*}[t]
    \centering
    \small 
    \setlength{\tabcolsep}{1mm}
    \centering

  \tablestyle{4pt}{1.2}\scriptsize\begin{tabular}{l| l|cc | x{15}x{15}x{15}x{15}x{35}x{15}x{40}}
  method & backbone & \# real image & \# synthetic image & \textbf{$\delta_1$}$\uparrow$ & \textbf{$\delta_2$}$\uparrow$ & \textbf{$\delta_3$}$\uparrow$ & REL~$\downarrow$ & Sq REL~$\downarrow$ & RMS~$\downarrow$ & RMS log~$\downarrow$  \\
  \shline
  Baseline & Swin-L & 50 & - & 0.59 & 0.84 & 0.93 & 0.31 & 0.37 & 0.81  &0.30\\
  \textbf{\modelname}  &  Swin-L & - & 35k~(R:50) & 0.68 & 0.90 & 0.97 & 0.22 & 0.19 & 0.60  & 0.23\\
  \textbf{\modelname}  &  Swin-L & 50 & 35k~(R:50)&  0.68 & 0.91 & 0.98 & 0.21 & 0.18 & 0.63  & 0.23\\
  \hline
  Baseline & Swin-L & 250 & - & 0.79 & 0.96 & 0.99 & 0.16 & 0.11 & 0.51  & 0.19\\
  \textbf{\modelname}  &  Swin-L & - & 35k~(R:250) & 0.78 & 0.96 & 0.99 & 0.17 & 0.11 & 0.52  & 0.19\\
  \textbf{\modelname}  &  Swin-L & 250 & 35k~(R:250)& 0.80  & 0.97 & 0.99 & 0.14 & 0.09 & 0.47  & 0.18\\
  \end{tabular}
    % %%%\vspace{-0.2cm}
    \caption{\textbf{Depth Estimation on NYU Depth V2 \texttt{val} dataset.}  Measurements are made for the depth range from $0m$ to $10m$.}
    % %%%\vspace{+2mm}
    \label{NYU_depth}
    \vspace{-2mm}
\end{table*}

\begin{table*}[t]
    \centering
    \small 
    \setlength{\tabcolsep}{1mm}
    \tablestyle{4pt}{1.2}\scriptsize\begin{tabular}{l|c|ccc|ccc}
    &  & \multicolumn{3}{c|}{Train Set/\%} &  \multicolumn{3}{c}{mIoU/\%}  \\
  methods & backbone & \# real image & \# synthetic image  & categories &  seen & unseen & harmonic \\
  \shline
  % \textit{Real Data} & & & & &  &   \\
  % \multicolumn{6}{l}{\textit{Manual \textbf{\underline{Mask}} Supervision}}\\
  ZS3~\cite{bucher2019zero}   & -  & 10.6k  & - & 15  & 78.0        & 21.2   & 33.3     \\
  CaGNet~\cite{gu2020context}  & -  & 10.6k  & - & 15  & 78.6        & 30.3   & 43.7     \\
  Joint~\cite{baek2021exploiting}  & -    &  10.6k  & -  &  15  & 77.7        & 32.5   & 45.9     \\
  STRICT~\cite{pastore2021closer}   & -   & 10.6k  & - &   15   & 82.7        & 35.6   & 49.8     \\
  SIGN~\cite{cheng2021sign}      & -      &10.6k  & - & 15  &  83.5       & 41.3   & 55.3     \\
  ZegFormer \cite{ding2022decoupling} & -  &10.6k & - & 15 &   86.4        & 63.6   & 73.3 \\

  \shline
  Li \textit{et al.}~\cite{li2023guiding} & ResNet101 &  & 10.0k~(R:110k, COCO) &15+5& 62.8   &    50.0    &  55.7 \\
  DiffuMask~\cite{wu2023diffumask}  & ResNet101& -  & 200.0 k~(R:0) & 15+5 & 62.1 & 50.5 & 55.7   \\
  DiffuMask~\cite{wu2023diffumask} & Swin-B & - & 200.0k~(R:0) & 15+5 & 71.4 & 65.0 & 68.1  \\
  \modelname & ResNet101 & - & 40k~(R:450, VOC) & 15+5 & 65.1 & 51.1  & 57.1\\ 
  \modelname & Swin-B & - & 40k~(R:450, VOC) & 15+5 & 78.8 & 60.5  & 68.4
  
  \end{tabular}

%   classes          IoU      nIoU
% --------------------------------
% road          : 0.984      nan
% sidewalk      : 0.868      nan
% building      : 0.927      nan
% wall          : 0.500      nan
% fence         : 0.607      nan
% pole          : 0.691      nan
% traffic light : 0.660      nan
% traffic sign  : 0.768      nan
% vegetation    : 0.926      nan
% terrain       : 0.650      nan
% sky           : 0.949      nan
% person        : 0.830    0.652
% rider         : 0.592    0.348
% car           : 0.945    0.875
% truck         : 0.645    0.334
% bus           : 0.919    0.641
% train         : 0.811    0.541
% motorcycle    : 0.676    0.352
% bicycle       : 0.781    0.566
% --------------------------------
% Score Average : 0.775    0.539
    % %%%\vspace{-0.2cm}
    \caption{\textbf{Zero-Shot Semantic Segmentation on PASCAL VOC 2012.} `Seen', `Unseen', and `Harmonic' denote the mIoU of seen, and unseen categories, and their harmonic mean. `R: ' refers to the number of real data from VOC 2012 or COCO 2017 used to train the generation model.}
    % %%%\vspace{+2mm}
    \label{VOC_open}
    \vspace{-2mm}
\end{table*}

\subsection{Baseline for Downstream Tasks}
\begin{itemize}
     \itemsep 0.0cm 
    \item \textbf{Semantic/Instance Segmentation.} We use Mask2former~\cite{cheng2022masked} as the baseline for comparing synthetic data to real data.
    We %employ
    use 
    the official code\footnote{\tt https://github.com/facebookresearch/MaskFormer}, maintaining all network settings, loss functions, and configurations as presented in the original code.
    To evaluate the effectiveness of synthetic data, we establish three settings: 1) training with purely real data, 2) training with purely synthetic data, and 3) joint training with both synthetic and real data.
    
    \item \textbf{Open-Vocabulary Semantic Segmentation.} Similar to the generic semantic segmentation, we use Mask2former~\cite{cheng2022masked} as the baseline. 
    We train DatasetDM on 15 seen categories and generate a total of 40k synthetic images for 20 categories. 
    Subsequently, we utilize these data to train the Mask2former model and evaluate its performance on the 20 categories of VOC 2012.

    \item \textbf{Depth Estimation.} DepthFormer~\cite{li2022depthformer}\footnote{\tt https://github.com/zhyever/Monocular-Depth-Estimation-Toolbox}, serving as the baseline, is employed to assess our approach.
    We adhere to all network settings, loss functions, configurations, and training strategies outlined in the original implementation.

    \item \textbf{Pose Estimation.} We adopt HRNet~\cite{sun2019deep} and its official code\footnote{\tt https://github.com/HRNet/HRNet-Human-Pose-Estimation} for evaluating the pose estimation task on synthetic data generated by \modelname. Currently, we focus on single-person scenarios in each synthetic image and guide GPT-4 to generate corresponding images accordingly.

\end{itemize}

\subsection{Training Setup for \modelname}

All experiments for training \modelname were carried out on a single V100 GPU, while downstream task baselines~(\ie{} Mask2former, Depthformer) were trained using 8 V100 GPUs.
Training our \modelname for $50k$ iterations with just one V100 GPU takes merely a day, showcasing its efficacy and efficiency.
For all tasks, we employ the Adam optimizer~\cite{loshchilov2017decoupled} with a learning rate of 0.001 and a batch size of 1.
The loss function and data augmentations vary for different tasks.

\begin{itemize}
     \itemsep 0.0cm 
    \item \textbf{Semantic/Instance Segmentation.} 
    During the training phase of \modelname, we primarily utilize two data augmentation techniques: random cropping to a size of 512$\times$512 pixels, and random scaling.
    \item \textbf{Depth Estimation.} 
    For depth estimation, we employ four data augmentation methods: random flipping, cropping, brightness-contrast adjustment, and hue-saturation value manipulation.
    \item \textbf{Pose Estimation.} 
    For pose estimation, we use four data augmentation techniques: random scaling, cropping, flipping, and rotation.
\end{itemize}

\subsection{Details for Training Data of \modelname}

\textbf{Quantities of training read data.}
Table   \ref{data} provides a comprehensive comparison of the quantities of training read data and synthetic data used for each downstream task in this study. Notably, with the exception of the seen class in the zero-shot segmentation setting, training with our \modelname requires less than $1\%$ of the available real data. This efficiency potentially reduces the implementation costs of perception algorithms and significantly improves data utilization.

\begin{table*}[t]
    \centering
    \small 
    \setlength{\tabcolsep}{1mm}
    \tablestyle{4pt}{1.2}\scriptsize\begin{tabular}{l|c|c|c|c|c|c}
    &  & & \multicolumn{3}{c|}{Train Set/\%} &    \\
  methods & baseline & backbone & \# labeled real image & \# unlabeled synthetic image  & \# synthetic image &  mIoU \\
  \shline
  ReCo~\cite{liu2021bootstrapping}   & DeepLabv3  & R101  & 60  & 10.6k-60  & 0        & 53.3     \\
  ReCo~\cite{liu2021bootstrapping}   & DeepLabv3  & R101  & 200  & 10.6k-200  & 0        & 69.8     \\
  ReCo~\cite{liu2021bootstrapping}   & DeepLabv3  & R101  & 600  & 10.6k-600  & 0        & 72.8     \\
  \modelname & DeepLabv3  & R101  & 60  & 0  & 40k        & 57.6     \\
  \modelname & Mask2former  & R50  & 100  & 0  & 40k        & 66.1     \\
  \modelname & Mask2former  & Swin-B  & 100  & 0  & 40k        & 78.5 
  \end{tabular}

    % %%%\vspace{-0.2cm}
    \caption{\textbf{Comparisons with semi-supervised works on PASCAL VOC 2012.}}
    % %%%\vspace{+2mm}
    \label{sim_supverision}
    \vspace{-2mm}
\end{table*}

\section{Experiments}

\subsection{Comparison with Other Data Augmentation Methods.}
From a certain perspective, the proposed \modelname is more akin to an efficient data augmentation technique, and thus we compare it with some previous data augmentation schemes, as shown in
Table   \ref{ablation_aug_baseline}.
Compared with flip and color augmentation, \modelname demonstrates a substantial advantage, bringing  significant improvements, around $10\%$ increase, which is significant for the computer vision community.

\subsection{Ablation Study for Baseline of Downstream Tasks.}
In addition, the synthetic data generated by \modelname can seamlessly integrate with any existing downstream task model.
To substantiate this claim, we tested our model with several other benchmark models, such as DeepLabV3, the results of which are detailed in Table   \ref{ablation_aug_baseline}. 
Notably, our synthetic data was able to enhance the performance of DeepLabV3 by approximately $10\%$, underscoring the robustness of our approach.

\subsection{Human Pose Estimation on COCO \texttt{val2017}.}
Table   \ref{COCO_Pos} presents the results of human pose estimation on the COCO 2017 dataset. 
Following the approach of HRNet~\cite{sun2019deep}, we established three distinct experimental settings, including variations in the backbone and input size, to evaluate the synthetic data from our model. 
Irrespective of the specific setting, our method consistently achieved an improvement of approximately $5\%$ in Average Precision (AP), which is a significant increase. Finally, it is noteworthy that our model attained competitive performance, with an AP of $48.9\%$, using merely 800 training images.

\subsection{Semantic segmentation on DeepFashion-MM.}
Table   \ref{deepfashion_sem} showcases the performance of semantic segmentation on the DeepFashion-MM dataset~\cite{jiang2022text2human}.
Like our other experiments, we have conducted two sets of experiments using different backbones.
Regardless of the setup, the joint training with synthetic data consistently outperforms the baseline that uses purely synthetic data, with an approximate improvement of $5\%$ mIoU.

\subsection{Depth Estimation on NYU Depth V2 val dataset.}
Table   \ref{NYU_depth} presents the depth estimation experiment conducted on the NYU Depth V2 validation dataset~\cite{silberman2012indoor}. 
Two training strategies have been devised based on variations in the training data.
Independent of the data volume utilized, our approach consistently yields substantial improvements, specifically $0.1$ and $0.02$ respectively.

\subsection{Zero-Shot Semantic Segmentation on VOC 2012}
Consistent with preceding studies~\cite{li2023guiding,wu2023diffumask}, we conduct an experiment on zero-shot (open-vocabulary) semantic segmentation tasks using the VOC 2012 dataset~\cite{(voc)everingham2010pascal}. 
Table   \ref{VOC_open} offers a comparative analysis with existing approaches to zero-shot semantic segmentation. 
In this experiment, our model is trained on a mere 450 images, with 30 images allocated for each of the 15 seen classes, and testing is conducted across all 20 categories.
Despite the limited dataset in comparison to the complete set of 10.6k images, our model continues to exhibit competitive performance. 
In relation to methods employing synthetic data, our model achieves state-of-the-art (SOTA) performance, reaching $68.4\%$ mIoU.

\subsection{Domain Generalization across Different Domains}
Following DiffuMask~\cite{wu2023diffumask}, we further assess the domain generalization capabilities of synthetic data produced by \modelname, as depicted in Fig. \ref{dg}. 
When compared with the previous state-of-the-art (SOTA) method, \modelname demonstrates superior effectiveness in domain generalization. 
For instance, \modelname achieves a score of $73.6\%$, as opposed to DiffuMask's score of $69.5\%$ on the VOC 2012 \texttt{val} set.
Compared to real data, \modelname exhibits enhanced robustness in terms of generalization.
It is reasonable that synthetic data exhibits greater diversity, especially when integrated with language models, as shown in Fig. \ref{fig:GPT}
In terms of diversity and robustness, it far surpasses real datasets.

\subsection{Comparison with the semi-supervised approaches on VOC2012}
Table~\ref{sim_supverision} presents the comparison between the prior semi-supervised works and our \modelname.
Even with a smaller amount of data (60 images), our approach demonstrates competitive performance, outperforming current semi-supervised semantic segmentation works.
Furthermore, with a more powerful backbone, our method can achieve even better performance, reaching a mIoU of 78.2 with only 100 images.

\begin{table}[t]
    \centering
    \small 
    \setlength{\tabcolsep}{1mm}
    \tablestyle{4pt}{1.2}\scriptsize\begin{tabular}{ r |c|ccc|c}
   & &  \multicolumn{4}{c}{mIoU/\%}  \\
  Train Set &  Test Set & Car & Person & Motorbike & mIoU\\
  \shline
  % \textit{Real Data} & & & & &  &   \\
  Cityscapes~\cite{cordts2016cityscapes} & VOC 2012~\cite{(voc)everingham2010pascal} \texttt{val}&  26.4 &  32.9  & 28.3    &     29.2    \\
  ADE20K~\cite{(ade)zhou2017scene}      & VOC 2012~\cite{(voc)everingham2010pascal} \texttt{val} &  73.2 & 66.6        & 64.1       & 68.0\\
  DiffuMask~\cite{wu2023diffumask}     & VOC 2012~\cite{(voc)everingham2010pascal} \texttt{val} &  74.2 &    71.0     & 63.2      & 69.5 \\
  \modelname     & VOC 2012~\cite{(voc)everingham2010pascal} \texttt{val} &   77.9 &   72.9 & 70.1 &  73.6\\
  \shline
  VOC 2012~\cite{(voc)everingham2010pascal}     &  Cityscapes~\cite{cordts2016cityscapes} \texttt{val}&   85.6       &  53.2    &  11.9 & 50.2\\
  ADE20K~\cite{(ade)zhou2017scene}       & Cityscapes~\cite{cordts2016cityscapes} \texttt{val}&    83.3       &  63.4  & 33.7 & 60.1 \\
  DiffuMask~\cite{wu2023diffumask}       & Cityscapes~\cite{cordts2016cityscapes} \texttt{val}& 84.0 &   70.7    &  23.6 & 59.4\\
  \modelname       & Cityscapes~\cite{cordts2016cityscapes} \texttt{val}&    85.6  &  58.9 & 12.7 & 52.4

  \end{tabular}

%   classes          IoU      nIoU
% --------------------------------
% road          : 0.984      nan
% sidewalk      : 0.868      nan
% building      : 0.927      nan
% wall          : 0.500      nan
% fence         : 0.607      nan
% pole          : 0.691      nan
% traffic light : 0.660      nan
% traffic sign  : 0.768      nan
% vegetation    : 0.926      nan
% terrain       : 0.650      nan
% sky           : 0.949      nan
% person        : 0.830    0.652
% rider         : 0.592    0.348
% car           : 0.945    0.875
% truck         : 0.645    0.334
% bus           : 0.919    0.641
% train         : 0.811    0.541
% motorcycle    : 0.676    0.352
% bicycle       : 0.781    0.566
% --------------------------------
% Score Average : 0.775    0.539
    % %%%\vspace{-0.2cm}
    \caption{\textbf{Performance for Domain Generalization between different datasets.} Mask2former~\cite{cheng2022masked} with ResNet50 is used as the baseline. \texttt{Person} and \texttt{Rider} classes of Cityscapes~\cite{cordts2016cityscapes} are consider as the same class, \ie{}, \texttt{Person} in the experiment. }
    % %%%\vspace{+2mm}
    % %%%\vspace{+2mm
    \label{dg}
    % %%%\vspace{-2mm}
\end{table}

\begin{figure}
    \centering
    \includegraphics[width=\linewidth]{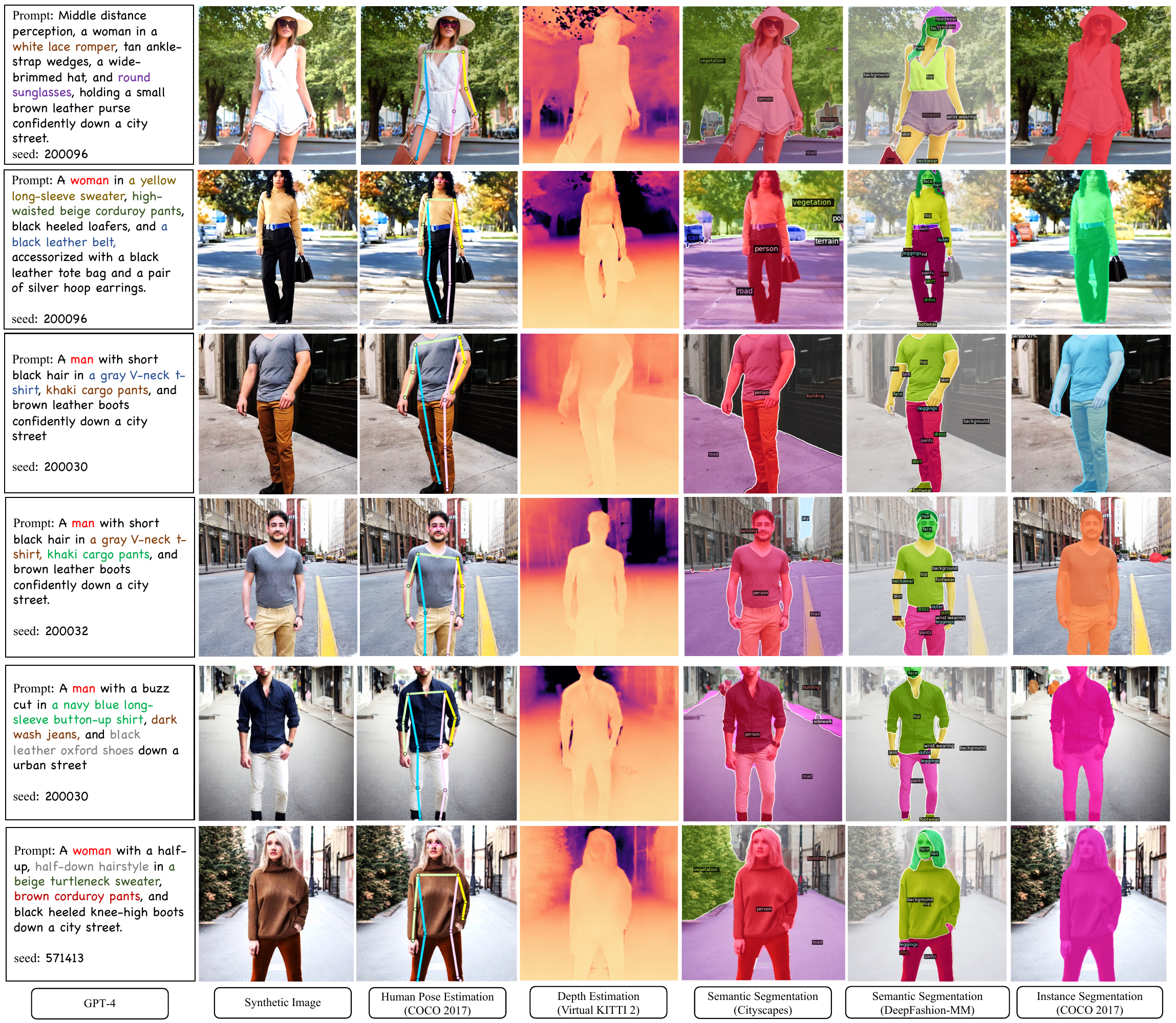}
    %%%\vspace{-6mm}
    \caption{
        \textbf{ Examples of Human-Centric Generated Data for \modelname.} Our method is capable of generating high-accuracy, high-diversity, and unified perceptual annotations.
        }
    \label{fig:Examples_human}
    %%%\vspace{-1mm}
\end{figure}

\subsection{More Qualitative Results}
\begin{figure}[t]
    \centering
    \includegraphics[width=\linewidth]{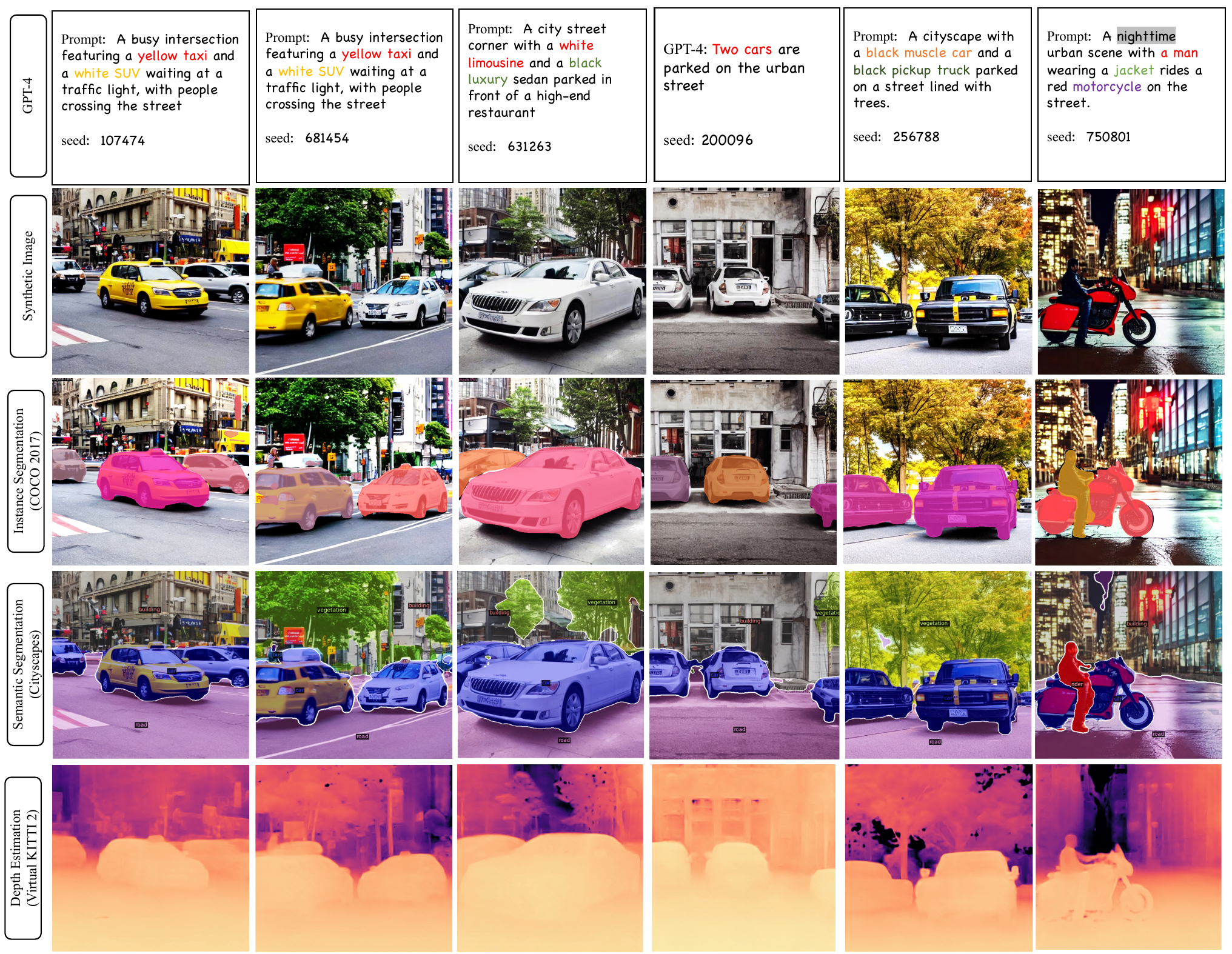}
    %%%\vspace{-6mm}
    \caption{
        \textbf{ Examples of  Generated Data for Urban City Scenario from \modelname.}
        }
    \label{fig:Examples_city}
    %%%\vspace{-2mm}
\end{figure}

To demonstrate the high-quality synthetic data, we visualized synthetic data from two domains: human-centric and urban city, as shown in Fig. \ref{fig:Examples_human} (human-centric) and Fig. \ref{fig:Examples_city} (urban city scenario).
The human-centric domain predominantly encompasses datasets related to human activity, such as COCO 2017, Cityscapes, and DeepFashion-MM. On the other hand, the urban city scenario pertains specifically to datasets like Cityscapes and COCO 2017.
To the best of our knowledge, our work is the first to support multi-task synthesis of data.
We believe that unified annotation synthesis is meaningful and can support interactions between different modalities. 
Recent works, \eg{} ImageBind~\cite{girdhar2023imagebind} have already demonstrated its feasibility and necessity.
Our method also has many advantages, such as the ability to custom design datasets for a specific domain or to address bad case scenarios, and it is particularly effective in solving problems related to long-tail data distribution.
This is straightforward; we can achieve it simply by adjusting our prompts.

\begin{figure}[t]
    \centering
    \includegraphics[width=\linewidth]{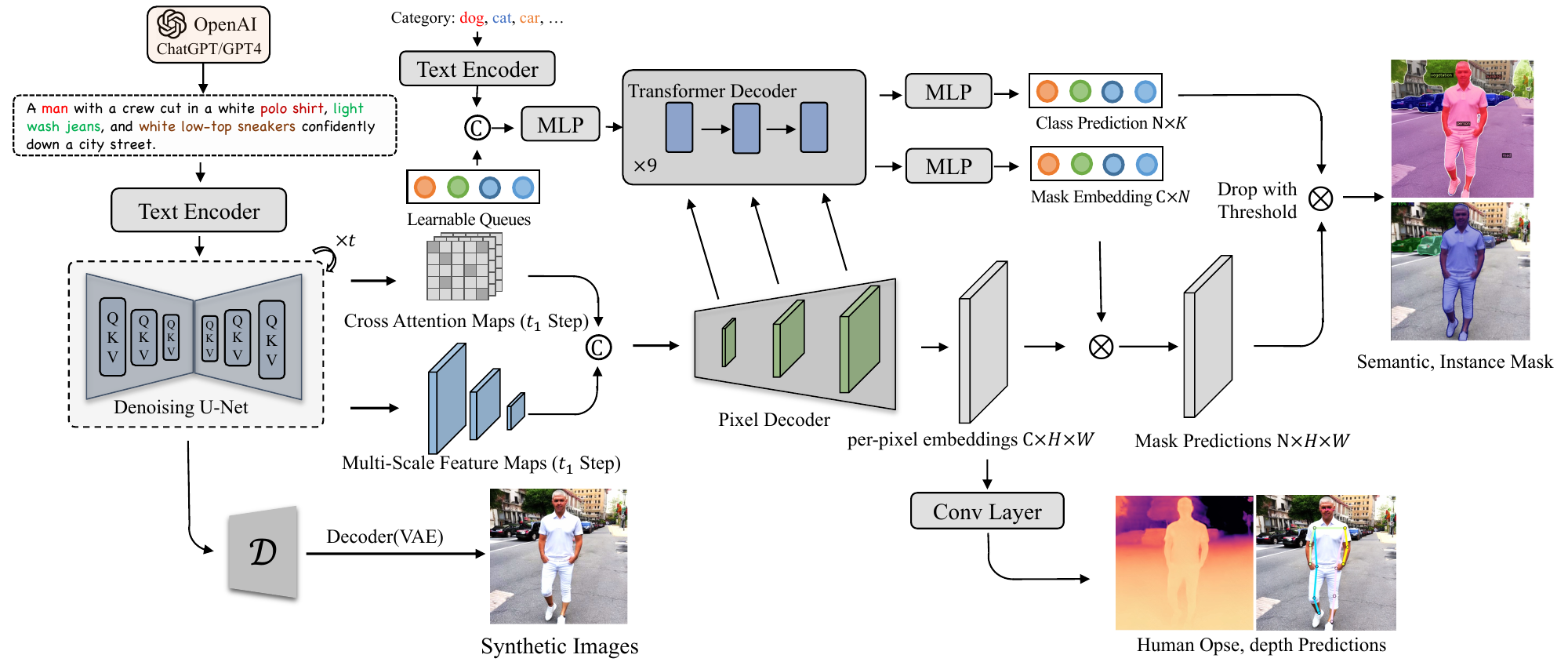}
    %%%\vspace{-6mm}
    \caption{
        \textbf{Details for P-Decoder.}  The whole framework of decoder  consists of text encoder, pixel decoder, and transformer decoder. For different downstream task, we  only need to adjust \textit{minor variations} \ie{} whether to startup some layers.
        }
    \label{fig:decoder}
    %%%\vspace{-1mm}
\end{figure}

\section{Details on the Architecture of Perception Decoder}
We show the detailed architecture of our P-Decoder in Fig. \ref{fig:decoder},
which consists of pixel decoder, text encoder, transformer decoder.

\subsection{Text Encoder for Open-Vocabulary Segmentation.}
In the open-vocabulary setting, for each class, we encode the corresponding class name~(\ie{} \textit{cat}, \textit{dog}) into a d-dimensional vector using the CLIP encoder. 
For a word corresponding to two text tokens, we average them into one token.
Subsequently, this token is replicated $n$ times, resulting in an $n \times d$ matrix. 
The matrix is then concatenated with a learnable query embedding of dimensions $n \times 768$. 
Ultimately, the concatenation is processed through a Multilayer Perceptron (MLP) layer to fuse the elements.

\subsection{Semantic and Instance Segmentation.}
With the representation $\mathcal{\hat{F}}$, which is fused from multi-scale features and cross-attention maps, we employ a pixel decoder and a transformer decoder to derive the per-pixel embedding $C\times H \times W$ and mask embedding $C \times N$.
As per the method outlined by Li.\textit{et al}~\cite{li2023guiding}, the pixel decoder consists of several straightforward up-sampling layers. Each layer comprises four types of computations: 
1) \texttt{$1\times1 $ Conv} for adjusting feature dimensionality, 
2) \texttt{Upsample} using simple linear interpolation to upscale the feature to a higher spatial resolution, 
3) Concat for merging features from different layers, and 
4) \texttt{Mix-conv} for blending features from varying spatial resolutions, which includes two $3\times3$ Conv.
Similar to Mask2former~\cite{cheng2022masked}, the transformer decoder comprises a stack of transformer layers with cross-attention, self-attention, and masked attention.
The final mask predictions of dimensions $N \times H \times W$ can be obtained by performing a simple matrix multiplication of the per-pixel embedding of dimensions $C \times H \times W$ and the mask embedding of dimensions $C \times N$.

\subsection{Human Pose and Depth Estimation.}
By expanding the segmentation architecture with the addition of two convolutional layers to the pixel decoder, we are able to efficiently handle the associated tasks of pose and depth estimation. 
Consequently, we derive two predictive outputs, denoted by $\mathbf{O} \in \mathcal{1}^{M\times H \times W}$ and $\mathbf{O} \in \mathcal{17}^{M\times H \times W}$, corresponding to the human pose and depth estimation tasks, respectively.

\begin{table}[t]
    \centering
    \small 
    \setlength{\tabcolsep}{1mm}
    \tablestyle{4pt}{1.2}\scriptsize\begin{tabular}{ l |c|p{0.6\columnwidth}}
  Dataset &  Category & Prompts\\
  \shline
  \multirow{6}*{VOC 2012~\cite{(voc)everingham2010pascal}}  & \multirow{2}*{Car} & A classic red convertible parked near a sandy beach, its vibrant color contrasting with the clear blue sky.\\
    &  & ...\\ 
    \cline{2-3}
   & \multirow{2}*{Person} & A young woman jogging in a park, wearing athletic clothing and listening to music through her earphones.\\ 
     &  & ...\\ 
     \cline{2-3}
       & \multirow{2}*{Dog} & A playful Golden Retriever, its fur gleaming in the sunlight, splashing in the water at a dog-friendly beach.\\ 
         &  & ...\\ 
  \shline
  \multirow{6}*{Cityscapes~\cite{cordts2016cityscapes}}  & \multirow{2}*{Car} & A sleek, black car cruises down a busy urban street lined with towering skyscrapers.\\
    &  & ...\\ 
    \cline{2-3}
   & \multirow{2}*{Person} & A woman in a red dress is crossing the street at the crosswalk while cars wait for her.\\ 
     &  & ...\\
    \cline{2-3}
       & \multirow{2}*{Bus} & A red double-decker bus drives through the heart of the city on a busy urban street, the passengers admiring the sights from the upper level.\\ 
         &  & ...\\ 
   \shline
  \multirow{6}*{COCO 2017~\cite{lin2014microsoft}}  & \multirow{2}*{Car} & Two red cars parked on a busy city street in the afternoon.\\
    &  & ...\\ 
    \cline{2-3}
   & \multirow{2}*{Person} & A group of people playing volleyball on a beach, with the ocean in the background.\\ 
     &  & ...\\
    \cline{2-3}
    & \multirow{2}*{Bus} & A big red bus parked on the side of the road with a tree behind it.\\ 
         &  &...\\ 
  \shline
  \multirow{3}*{NYU Depth V2~\cite{silberman2012indoor}}  &  & A kitchen with white cabinets, stainless steel appliances, and a wooden table.\\
    & - & A bedroom with a queen-size bed, dresser, and nightstand.\\ 
    &  & ...\\ 
   \shline
   \multirow{3}*{COCO 2017 Pose~\cite{lin2014microsoft}}  &  & a person with a backpack, wearing a green jacket and khaki pants.\\
    & - & a middle-aged woman wearing a red blazer, black slacks, and pumps.\\ 
    &  & ...\\ 
   % \shline
   % \multirow{3}*{NYU Depth V2~\cite{silberman2012indoor}}  &  & \\
   %  & - & \\ 
   %  &  & \\ 
   \shline
   \multirow{3}*{DeepFashion-MM~\cite{jiang2022text2human}}  &  & A woman wearing a loose-fitting white blouse with ruffled sleeves, paired with high-waisted, wide-leg navy blue pants and black ankle-strap stiletto heels.\\
    & - & A man dressed in a classic white button-up shirt, khaki chinos with a slim fit, and brown suede desert boots.\\ 
    &  & ...\\ 
  \end{tabular}

    \caption{\textbf{Prompts for Different Datasets.} For different data domains, we will guide GPT-4 to generate corresponding prompts. We will release the code and corresponding prompts files.}

    \label{prompt}
    \vspace{-0.2cm}
\end{table}

\section{Synthetic Dataset}

\subsection{Prompts from GPT-4}
\begin{figure}[t]
    \centering
    \includegraphics[width=\linewidth]{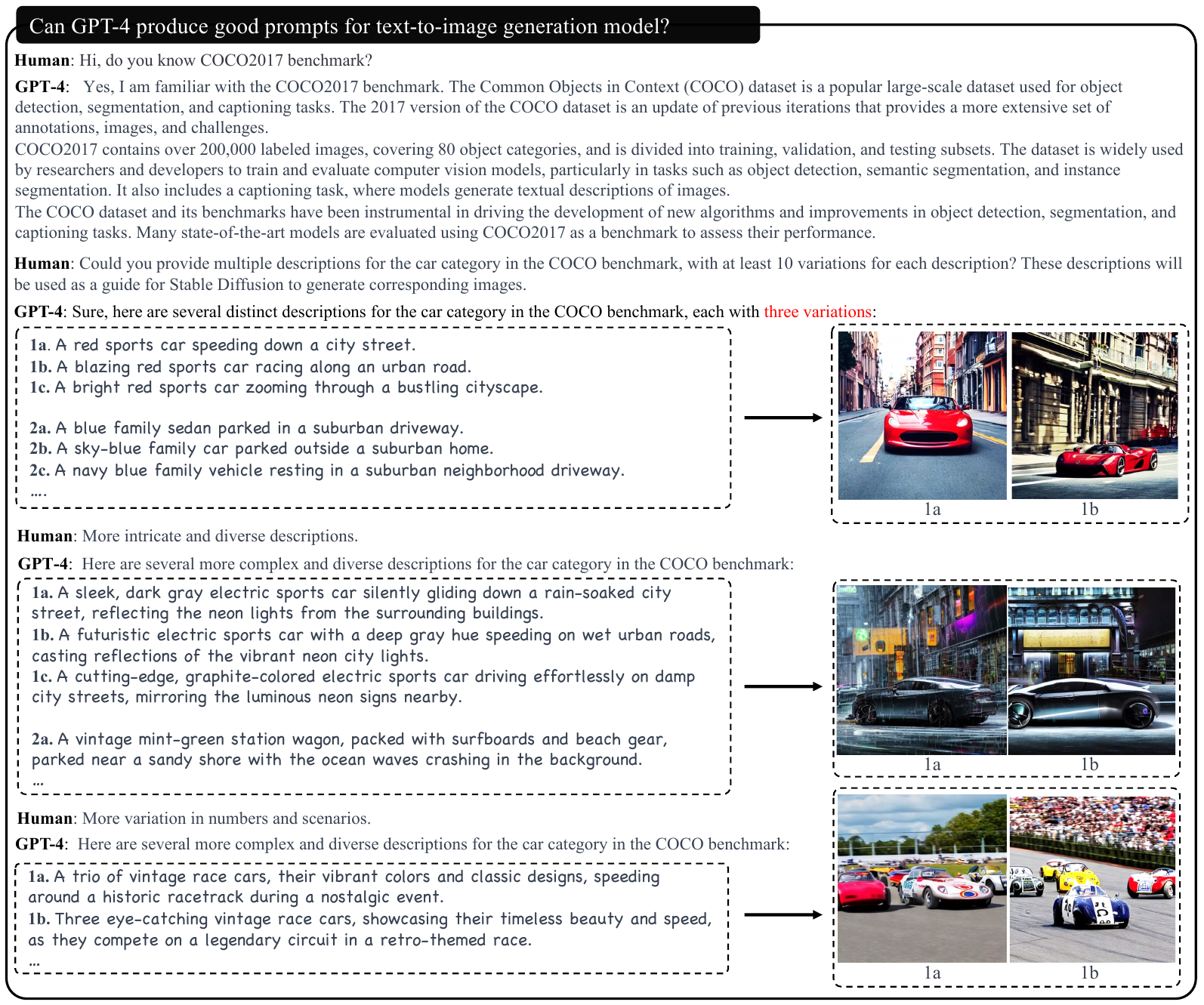}
    %%%\vspace{-5mm}
    \caption{
        \textbf{ Prompts of diffusion model from GPT-4.} By providing some simple cues, GPT-4 can generate a vast and diverse array of prompts. 
        }
    \label{fig:GPT}
    \vspace{-2mm}
\end{figure}

Here, we also demonstrate the detailed process of prompt generation, guided by GPT-4, as shown in Fig. \ref{fig:GPT}.
Throughout the process, human only need to provide a small number of prompts to guide GPT-4. 
With a cost of no more than 50 words prompt clue, we can accomplish the generation of a massive number of prompts for a downstream task dataset.
It is worth mention that text-guided data is extremely flexible. 
We can customize the generation of certain attributes of data domain.
For instance, if we need to enhance the variation in the number of objects, we can provide a prompt like \texttt{More variation in number}. 
This is extremely flexible and convenient.

\subsection{Prompts for Each Dataset}
As shown in Table   \ref{prompt}, we also provide some prompt cases of our method for each dataset, and we will open-source these prompts along with the corresponding code.
For tasks that distinguish between classes, \ie{} semantic and instance segmentation, we will guide GPT-4 to generate around 100 descriptions specifically for each class. 
For tasks and datasets that are not class-sensitive, \eg{} pose and depth estimation, we guide GPT-4 to generate a large number of descriptions all at once.

\section{Limitation and Future Work}

\subsection{Potential Negative Societal Impacts}
As with other projects involving synthetic image generation, the potential adverse societal implications of our work largely revolve around ethical considerations. 
Utilizing the Stable Diffusion model, trained on the 5-billion image LION dataset, raises notable private copyright concerns due to the nature of the dataset.
Nonetheless, we maintain that such potential repercussions should not overshadow the applicability and value of our research. Numerous text-guided text-to-image diffusion works exist, of which ours is but one. 
We also emphasize that it's feasible to mitigate these ethical issues through the careful crafting of specific prompts, serving as an effective countermeasure.
Additionally, we can mitigate this impact from an algorithmic perspective~\cite{kumari2023ablating}, such as eliminating certain concepts (which may infringe on personal privacy) from the pretrained model.

\begin{figure}[t]
    \centering
    \includegraphics[width=\linewidth]{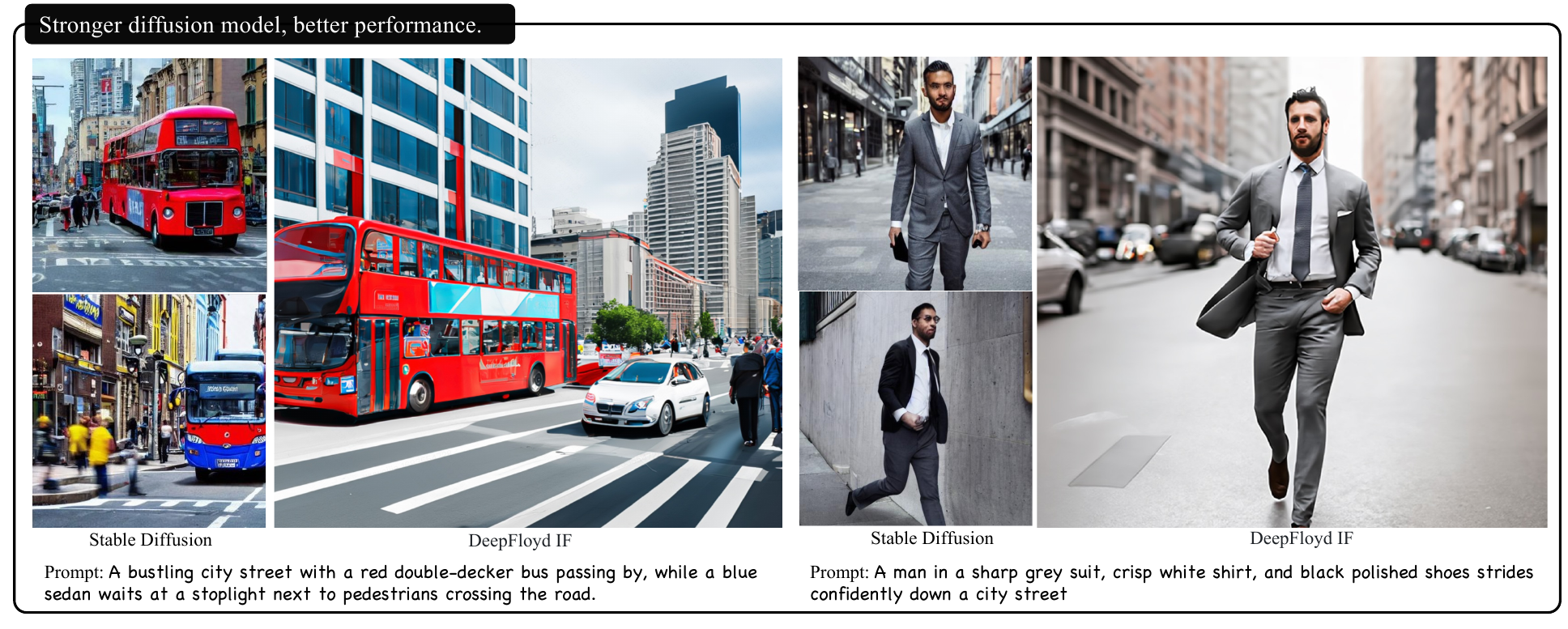}
    %%%\vspace{-5mm}
    \caption{
        \textbf{Stronger Diffusion Model, Greater Potential.}
        With the advancements in generative models, synthetic data will have greater potential and possibilities for perception tasks. A simple solution is to replace Stable Diffusion with DeepFolyd IF directly.}
    \label{fig:IF}
    \vspace{-1mm}
\end{figure}

\subsection{Limitation \& Future Work}
\textbf{Limitation.}
The main limitation of this study is that the quality and complexity of the synthesized data still cannot compare with real data. 
If certain companies and organizations could invest substantial resources to collect and manually annotate massive amounts of training data, better results could be achieved. 
However, this is actually the main limitation of the Stable diffusion model. We could also consider using more powerful diffusion models to alleviate this issue, as shown in Fig. \ref{fig:IF}.
Deepfloyd IF~\footnote{https://github.com/deep-floyd/IF} is a more powerful text-guided image generation model, which significantly outperforms Stable diffusion in two main aspects. 
First, it excels in semantic alignment - given a lengthy text description, the IF model can generate related images more accurately. 
Second, the IF model can synthesize images at a higher resolution, specifically 1024, while the resolution of Stable diffusion is only 512. 
We believe that our method, in combination with the DeepFloyd IF model, can lead to further improvements and make a greater contribution. 
Due to time constraints~(DeepFloyd IF released on May 2023), we are unable to provide related experiments, but this does not affect the validation of the effectiveness of our method. 
Our primary contribution lies in using a unified decoder to parse the latent space of the pre-trained diffusion model, not in enhancing the quality of image synthesis.

\textbf{Future Work}
This study is intriguing and innovative, possessing profound exploratory significance. 
We identify several avenues for future enhancement: firstly, employing a more robust text-guided image generation model may yield substantial improvements. 
Secondly, augmenting the efficiency of prompt generation, or designing prompts that better align with the target domain could prove beneficial. 
For example, synthesizing specific prompts corresponding to the COCO 2017 dataset could be viable.

% \small
% \bibliographystyle{abbrv}
% % \bibliographystyle{abbrv}
% % \bibliographystyle{unsrtnat}
% % \bibliographystyle{alpha}
% \bibliography{sample}

\bibliographystyle{abbrv}
{
\small
\bibliography{neurips_2023}

\begin{thebibliography}{10}

\bibitem{ahn2018learning}
J.~Ahn and S.~Kwak.
\newblock Learning pixel-level semantic affinity with image-level supervision
  for weakly supervised semantic segmentation.
\newblock In {\em Proceedings of the IEEE conference on computer vision and
  pattern recognition}, pages 4981--4990, 2018.

\bibitem{akiva2021towards}
P.~Akiva and K.~Dana.
\newblock Towards single stage weakly supervised semantic segmentation.
\newblock {\em arXiv preprint arXiv:2106.10309}, 2021.

\bibitem{baek2021exploiting}
D.~Baek, Y.~Oh, and B.~Ham.
\newblock Exploiting a joint embedding space for generalized zero-shot semantic
  segmentation.
\newblock In {\em Proc. ICCV}, 2021.

\bibitem{baranchuk2021label}
D.~Baranchuk, I.~Rubachev, A.~Voynov, V.~Khrulkov, and A.~Babenko.
\newblock Label-efficient semantic segmentation with diffusion models.
\newblock {\em arXiv preprint arXiv:2112.03126}, 2021.

\bibitem{bubeck2023sparks}
S.~Bubeck, V.~Chandrasekaran, R.~Eldan, J.~Gehrke, E.~Horvitz, E.~Kamar,
  P.~Lee, Y.~T. Lee, Y.~Li, S.~Lundberg, et~al.
\newblock Sparks of artificial general intelligence: Early experiments with
  gpt-4.
\newblock {\em arXiv preprint arXiv:2303.12712}, 2023.

\bibitem{bucher2019zero}
M.~Bucher, T.-H. Vu, M.~Cord, and P.~P{\'e}rez.
\newblock Zero-shot semantic segmentation.
\newblock {\em NeurIPS}, 2019.

\bibitem{butler2012naturalistic}
D.~J. Butler, J.~Wulff, G.~B. Stanley, and M.~J. Black.
\newblock A naturalistic open source movie for optical flow evaluation.
\newblock In {\em Computer Vision--ECCV 2012: 12th European Conference on
  Computer Vision, Florence, Italy, October 7-13, 2012, Proceedings, Part VI
  12}, pages 611--625. Springer, 2012.

\bibitem{cabon2020virtual}
Y.~Cabon, N.~Murray, and M.~Humenberger.
\newblock Virtual kitti 2.
\newblock {\em arXiv preprint arXiv:2001.10773}, 2020.

\bibitem{chen2018encoder}
L.-C. Chen, Y.~Zhu, G.~Papandreou, F.~Schroff, and H.~Adam.
\newblock Encoder-decoder with atrous separable convolution for semantic image
  segmentation.
\newblock In {\em Proceedings of the European conference on computer vision
  (ECCV)}, pages 801--818, 2018.

\bibitem{cheng2022masked}
B.~Cheng, I.~Misra, A.~G. Schwing, A.~Kirillov, and R.~Girdhar.
\newblock Masked-attention mask transformer for universal image segmentation.
\newblock In {\em Proceedings of the IEEE/CVF Conference on Computer Vision and
  Pattern Recognition}, pages 1290--1299, 2022.

\bibitem{cheng2021per}
B.~Cheng, A.~Schwing, and A.~Kirillov.
\newblock Per-pixel classification is not all you need for semantic
  segmentation.
\newblock {\em Advances in Neural Information Processing Systems},
  34:17864--17875, 2021.

\bibitem{cheng2021sign}
J.~Cheng, S.~Nandi, P.~Natarajan, and W.~Abd-Almageed.
\newblock Sign: Spatial-information incorporated generative network for
  generalized zero-shot semantic segmentation.
\newblock In {\em Proc. ICCV}, 2021.

\bibitem{cordts2016cityscapes}
M.~Cordts, M.~Omran, S.~Ramos, T.~Rehfeld, M.~Enzweiler, R.~Benenson,
  U.~Franke, S.~Roth, and B.~Schiele.
\newblock The cityscapes dataset for semantic urban scene understanding.
\newblock In {\em Proceedings of the IEEE conference on computer vision and
  pattern recognition}, pages 3213--3223, 2016.

\bibitem{(cityscape)cordts2016cityscapes}
M.~Cordts, M.~Omran, S.~Ramos, T.~Rehfeld, M.~Enzweiler, R.~Benenson,
  U.~Franke, S.~Roth, and B.~Schiele.
\newblock The cityscapes dataset for semantic urban scene understanding.
\newblock In {\em Proceedings of the IEEE conference on computer vision and
  pattern recognition}, pages 3213--3223, 2016.

\bibitem{ding2022decoupling}
J.~Ding, N.~Xue, G.-S. Xia, and D.~Dai.
\newblock Decoupling zero-shot semantic segmentation.
\newblock In {\em Proc. CVPR}, 2022.

\bibitem{dinh2014nice}
L.~Dinh, D.~Krueger, and Y.~Bengio.
\newblock Nice: Non-linear independent components estimation.
\newblock {\em arXiv preprint arXiv:1410.8516}, 2014.

\bibitem{eigen2014depth}
D.~Eigen, C.~Puhrsch, and R.~Fergus.
\newblock Depth map prediction from a single image using a multi-scale deep
  network.
\newblock {\em Advances in neural information processing systems}, 27, 2014.

\bibitem{esser2021taming}
P.~Esser, R.~Rombach, and B.~Ommer.
\newblock Taming transformers for high-resolution image synthesis.
\newblock In {\em Proceedings of the IEEE/CVF conference on computer vision and
  pattern recognition}, pages 12873--12883, 2021.

\bibitem{(voc)everingham2010pascal}
M.~Everingham, L.~Van~Gool, C.~K. Williams, J.~Winn, and A.~Zisserman.
\newblock The pascal visual object classes (voc) challenge.
\newblock {\em International journal of computer vision}, 88(2):303--338, 2010.

\bibitem{floridi2020gpt}
L.~Floridi and M.~Chiriatti.
\newblock Gpt-3: Its nature, scope, limits, and consequences.
\newblock {\em Minds and Machines}, 30:681--694, 2020.

\bibitem{gaidon2016virtual}
A.~Gaidon, Q.~Wang, Y.~Cabon, and E.~Vig.
\newblock Virtual worlds as proxy for multi-object tracking analysis.
\newblock In {\em Proceedings of the IEEE conference on computer vision and
  pattern recognition}, pages 4340--4349, 2016.

\bibitem{ge2020pose}
Y.~Ge, J.~Zhao, and L.~Itti.
\newblock Pose augmentation: Class-agnostic object pose transformation for
  object recognition.
\newblock In {\em Computer Vision--ECCV 2020: 16th European Conference,
  Glasgow, UK, August 23--28, 2020, Proceedings, Part XXVIII 16}, pages
  138--155. Springer, 2020.

\bibitem{girdhar2023imagebind}
R.~Girdhar, A.~El-Nouby, Z.~Liu, M.~Singh, K.~V. Alwala, A.~Joulin, and
  I.~Misra.
\newblock Imagebind: One embedding space to bind them all.
\newblock In {\em CVPR}, 2023.

\bibitem{goodfellow2020generative}
I.~Goodfellow, J.~Pouget-Abadie, M.~Mirza, B.~Xu, D.~Warde-Farley, S.~Ozair,
  A.~Courville, and Y.~Bengio.
\newblock Generative adversarial networks.
\newblock {\em Communications of the ACM}, 63(11):139--144, 2020.

\bibitem{gu2023mix}
Y.~Gu, X.~Wang, J.~Z. Wu, Y.~Shi, Y.~Chen, Z.~Fan, W.~Xiao, R.~Zhao, S.~Chang,
  W.~Wu, et~al.
\newblock Mix-of-show: Decentralized low-rank adaptation for multi-concept
  customization of diffusion models.
\newblock {\em arXiv preprint arXiv:2305.18292}, 2023.

\bibitem{gu2020context}
Z.~Gu, S.~Zhou, L.~Niu, Z.~Zhao, and L.~Zhang.
\newblock Context-aware feature generation for zero-shot semantic segmentation.
\newblock In {\em ACM MM}, 2020.

\bibitem{han2023medgen3d}
K.~Han, Y.~Xiong, C.~You, P.~Khosravi, S.~Sun, X.~Yan, J.~Duncan, and X.~Xie.
\newblock Medgen3d: A deep generative framework for paired 3d image and mask
  generation.
\newblock {\em arXiv preprint arXiv:2304.04106}, 2023.

\bibitem{he2022synthetic}
R.~He, S.~Sun, X.~Yu, C.~Xue, W.~Zhang, P.~Torr, S.~Bai, and X.~Qi.
\newblock Is synthetic data from generative models ready for image recognition?
\newblock {\em arXiv preprint arXiv:2210.07574}, 2022.

\bibitem{hendy2023good}
A.~Hendy, M.~Abdelrehim, A.~Sharaf, V.~Raunak, M.~Gabr, H.~Matsushita, Y.~J.
  Kim, M.~Afify, and H.~H. Awadalla.
\newblock How good are gpt models at machine translation? a comprehensive
  evaluation.
\newblock {\em arXiv preprint arXiv:2302.09210}, 2023.

\bibitem{hertz2022prompt}
A.~Hertz, R.~Mokady, J.~Tenenbaum, K.~Aberman, Y.~Pritch, and D.~Cohen-Or.
\newblock Prompt-to-prompt image editing with cross attention control.
\newblock {\em arXiv preprint arXiv:2208.01626}, 2022.

\bibitem{ho2020denoising}
J.~Ho, A.~Jain, and P.~Abbeel.
\newblock Denoising diffusion probabilistic models.
\newblock {\em Advances in Neural Information Processing Systems},
  33:6840--6851, 2020.

\bibitem{ho2022cascaded}
J.~Ho, C.~Saharia, W.~Chan, D.~J. Fleet, M.~Norouzi, and T.~Salimans.
\newblock Cascaded diffusion models for high fidelity image generation.
\newblock {\em J. Mach. Learn. Res.}, 23(47):1--33, 2022.

\bibitem{jiang2022text2human}
Y.~Jiang, S.~Yang, H.~Qiu, W.~Wu, C.~C. Loy, and Z.~Liu.
\newblock Text2human: Text-driven controllable human image generation.
\newblock {\em ACM Transactions on Graphics (TOG)}, 41(4):1--11, 2022.

\bibitem{kingma2013auto}
D.~P. Kingma and M.~Welling.
\newblock Auto-encoding variational bayes.
\newblock {\em arXiv preprint arXiv:1312.6114}, 2013.

\bibitem{kumari2023ablating}
N.~Kumari, B.~Zhang, S.-Y. Wang, E.~Shechtman, R.~Zhang, and J.-Y. Zhu.
\newblock Ablating concepts in text-to-image diffusion models.
\newblock {\em arXiv preprint arXiv:2303.13516}, 2023.

\bibitem{li2022bigdatasetgan}
D.~Li, H.~Ling, S.~W. Kim, K.~Kreis, S.~Fidler, and A.~Torralba.
\newblock Bigdatasetgan: Synthesizing imagenet with pixel-wise annotations.
\newblock In {\em Proceedings of the IEEE/CVF Conference on Computer Vision and
  Pattern Recognition}, pages 21330--21340, 2022.

\bibitem{li2022depthformer}
Z.~Li, Z.~Chen, X.~Liu, and J.~Jiang.
\newblock Depthformer: Exploiting long-range correlation and local information
  for accurate monocular depth estimation.
\newblock {\em arXiv preprint arXiv:2203.14211}, 2022.

\bibitem{li2023guiding}
Z.~Li, Q.~Zhou, X.~Zhang, Y.~Zhang, Y.~Wang, and W.~Xie.
\newblock Guiding text-to-image diffusion model towards grounded generation.
\newblock {\em arXiv preprint arXiv:2301.05221}, 2023.

\bibitem{lin2016scribblesup}
D.~Lin, J.~Dai, J.~Jia, K.~He, and J.~Sun.
\newblock Scribblesup: Scribble-supervised convolutional networks for semantic
  segmentation.
\newblock In {\em Proceedings of the IEEE conference on computer vision and
  pattern recognition}, pages 3159--3167, 2016.

\bibitem{lin2014microsoft}
T.-Y. Lin, M.~Maire, S.~Belongie, J.~Hays, P.~Perona, D.~Ramanan,
  P.~Doll{\'a}r, and C.~L. Zitnick.
\newblock Microsoft coco: Common objects in context.
\newblock In {\em Computer Vision--ECCV 2014: 13th European Conference, Zurich,
  Switzerland, September 6-12, 2014, Proceedings, Part V 13}, pages 740--755.
  Springer, 2014.

\bibitem{lin2022raregan}
Z.~Lin, H.~Liang, G.~Fanti, and V.~Sekar.
\newblock Raregan: Generating samples for rare classes.
\newblock In {\em Proceedings of the AAAI Conference on Artificial
  Intelligence}, volume~36, pages 7506--7515, 2022.

\bibitem{ling2021editgan}
H.~Ling, K.~Kreis, D.~Li, S.~W. Kim, A.~Torralba, and S.~Fidler.
\newblock Editgan: High-precision semantic image editing.
\newblock {\em Advances in Neural Information Processing Systems},
  34:16331--16345, 2021.

\bibitem{liu2021bootstrapping}
S.~Liu, S.~Zhi, E.~Johns, and A.~J. Davison.
\newblock Bootstrapping semantic segmentation with regional contrast.
\newblock {\em arXiv preprint arXiv:2104.04465}, 2021.

\bibitem{loshchilov2017decoupled}
I.~Loshchilov and F.~Hutter.
\newblock Decoupled weight decay regularization.
\newblock {\em arXiv preprint arXiv:1711.05101}, 2017.

\bibitem{milletari2016v}
F.~Milletari, N.~Navab, and S.-A. Ahmadi.
\newblock V-net: Fully convolutional neural networks for volumetric medical
  image segmentation.
\newblock In {\em 2016 fourth international conference on 3D vision (3DV)},
  pages 565--571. IEEE, 2016.

\bibitem{pastore2021closer}
G.~Pastore, F.~Cermelli, Y.~Xian, M.~Mancini, Z.~Akata, and B.~Caputo.
\newblock A closer look at self-training for zero-label semantic segmentation.
\newblock In {\em Proc. CVPRW}, 2021.

\bibitem{pepik2012teaching}
B.~Pepik, M.~Stark, P.~Gehler, and B.~Schiele.
\newblock Teaching 3d geometry to deformable part models.
\newblock In {\em 2012 IEEE conference on computer vision and pattern
  recognition}, pages 3362--3369. IEEE, 2012.

\bibitem{radford2021learning}
A.~Radford, J.~W. Kim, C.~Hallacy, A.~Ramesh, G.~Goh, S.~Agarwal, G.~Sastry,
  A.~Askell, P.~Mishkin, J.~Clark, et~al.
\newblock Learning transferable visual models from natural language
  supervision.
\newblock In {\em International conference on machine learning}, pages
  8748--8763. PMLR, 2021.

\bibitem{ramesh2022hierarchical}
A.~Ramesh, P.~Dhariwal, A.~Nichol, C.~Chu, and M.~Chen.
\newblock Hierarchical text-conditional image generation with clip latents.
\newblock {\em arXiv preprint arXiv:2204.06125}, 2022.

\bibitem{rombach2022high}
R.~Rombach, A.~Blattmann, D.~Lorenz, P.~Esser, and B.~Ommer.
\newblock High-resolution image synthesis with latent diffusion models.
\newblock In {\em Proceedings of the IEEE/CVF Conference on Computer Vision and
  Pattern Recognition}, pages 10684--10695, 2022.

\bibitem{ronneberger2015u}
O.~Ronneberger, P.~Fischer, and T.~Brox.
\newblock U-net: Convolutional networks for biomedical image segmentation.
\newblock In {\em Medical Image Computing and Computer-Assisted
  Intervention--MICCAI 2015: 18th International Conference, Munich, Germany,
  October 5-9, 2015, Proceedings, Part III 18}, pages 234--241. Springer, 2015.

\bibitem{saharia2022photorealistic}
C.~Saharia, W.~Chan, S.~Saxena, L.~Li, J.~Whang, E.~L. Denton, K.~Ghasemipour,
  R.~Gontijo~Lopes, B.~Karagol~Ayan, T.~Salimans, et~al.
\newblock Photorealistic text-to-image diffusion models with deep language
  understanding.
\newblock {\em Advances in Neural Information Processing Systems},
  35:36479--36494, 2022.

\bibitem{satkin2012data}
S.~Satkin, J.~Lin, and M.~Hebert.
\newblock Data-driven scene understanding from 3d models.
\newblock 2012.

\bibitem{schuhmann2022laion}
C.~Schuhmann, R.~Beaumont, R.~Vencu, C.~Gordon, R.~Wightman, M.~Cherti,
  T.~Coombes, A.~Katta, C.~Mullis, M.~Wortsman, et~al.
\newblock Laion-5b: An open large-scale dataset for training next generation
  image-text models.
\newblock {\em arXiv preprint arXiv:2210.08402}, 2022.

\bibitem{silberman2012indoor}
N.~Silberman, D.~Hoiem, P.~Kohli, and R.~Fergus.
\newblock Indoor segmentation and support inference from rgbd images.
\newblock {\em ECCV (5)}, 7576:746--760, 2012.

\bibitem{sohl2015deep}
J.~Sohl-Dickstein, E.~Weiss, N.~Maheswaranathan, and S.~Ganguli.
\newblock Deep unsupervised learning using nonequilibrium thermodynamics.
\newblock In {\em International Conference on Machine Learning}, pages
  2256--2265. PMLR, 2015.

\bibitem{sun2019deep}
K.~Sun, B.~Xiao, D.~Liu, and J.~Wang.
\newblock Deep high-resolution representation learning for human pose
  estimation.
\newblock In {\em Proceedings of the IEEE/CVF conference on computer vision and
  pattern recognition}, pages 5693--5703, 2019.

\bibitem{wang2022deep}
X.~Wang, L.~Jing, Y.~Lyu, M.~Guo, J.~Wang, H.~Liu, J.~Yu, and T.~Zeng.
\newblock Deep generative mixture model for robust imbalance classification.
\newblock {\em IEEE Transactions on Pattern Analysis and Machine Intelligence},
  2022.

\bibitem{wang2022self}
Y.~Wang, Y.~Kordi, S.~Mishra, A.~Liu, N.~A. Smith, D.~Khashabi, and
  H.~Hajishirzi.
\newblock Self-instruct: Aligning language model with self generated
  instructions.
\newblock {\em arXiv preprint arXiv:2212.10560}, 2022.

\bibitem{wilson2020survey}
G.~Wilson and D.~J. Cook.
\newblock A survey of unsupervised deep domain adaptation.
\newblock {\em ACM Transactions on Intelligent Systems and Technology (TIST)},
  11(5):1--46, 2020.

\bibitem{wu2022medsegdiff}
J.~Wu, H.~Fang, Y.~Zhang, Y.~Yang, and Y.~Xu.
\newblock Medsegdiff: Medical image segmentation with diffusion probabilistic
  model.
\newblock {\em arXiv preprint arXiv:2211.00611}, 2022.

\bibitem{wu2023diffumask}
W.~Wu, Y.~Zhao, M.~Z. Shou, H.~Zhou, and C.~Shen.
\newblock Diffumask: Synthesizing images with pixel-level annotations for
  semantic segmentation using diffusion models.
\newblock {\em arXiv preprint arXiv:2303.11681}, 2023.

\bibitem{wu2022synthetic}
Z.~Wu, L.~Wang, W.~Wang, T.~Shi, C.~Chen, A.~Hao, and S.~Li.
\newblock Synthetic data supervised salient object detection.
\newblock In {\em Proceedings of the 30th ACM International Conference on
  Multimedia}, pages 5557--5565, 2022.

\bibitem{xu2022handsoff}
A.~Xu, M.~I. Vasileva, A.~Dave, and A.~Seshadri.
\newblock Handsoff: Labeled dataset generation with no additional human
  annotations.
\newblock {\em arXiv preprint arXiv:2212.12645}, 2022.

\bibitem{xu2023open}
J.~Xu, S.~Liu, A.~Vahdat, W.~Byeon, X.~Wang, and S.~De~Mello.
\newblock Open-vocabulary panoptic segmentation with text-to-image diffusion
  models.
\newblock {\em arXiv preprint arXiv:2303.04803}, 2023.

\bibitem{zhang2020survey}
M.~Zhang, Y.~Zhou, J.~Zhao, Y.~Man, B.~Liu, and R.~Yao.
\newblock A survey of semi-and weakly supervised semantic segmentation of
  images.
\newblock {\em Artificial Intelligence Review}, 53:4259--4288, 2020.

\bibitem{zhang2021datasetgan}
Y.~Zhang, H.~Ling, J.~Gao, K.~Yin, J.-F. Lafleche, A.~Barriuso, A.~Torralba,
  and S.~Fidler.
\newblock Datasetgan: Efficient labeled data factory with minimal human effort.
\newblock In {\em Proceedings of the IEEE/CVF Conference on Computer Vision and
  Pattern Recognition}, pages 10145--10155, 2021.

\bibitem{zhao2023unleashing}
W.~Zhao, Y.~Rao, Z.~Liu, B.~Liu, J.~Zhou, and J.~Lu.
\newblock Unleashing text-to-image diffusion models for visual perception.
\newblock {\em arXiv preprint arXiv:2303.02153}, 2023.

\bibitem{zhao2023generative}
Y.~Zhao, Q.~Ye, W.~Wu, C.~Shen, and F.~Wan.
\newblock Generative prompt model for weakly supervised object localization.
\newblock {\em arXiv preprint arXiv:2307.09756}, 2023.

\bibitem{(ade)zhou2017scene}
B.~Zhou, H.~Zhao, X.~Puig, S.~Fidler, A.~Barriuso, and A.~Torralba.
\newblock Scene parsing through ade20k dataset.
\newblock In {\em Proceedings of the IEEE conference on computer vision and
  pattern recognition}, pages 633--641, 2017.

\bibitem{zou2022generalized}
X.~Zou, Z.-Y. Dou, J.~Yang, Z.~Gan, L.~Li, C.~Li, X.~Dai, H.~Behl, J.~Wang,
  L.~Yuan, et~al.
\newblock Generalized decoding for pixel, image, and language.
\newblock {\em arXiv preprint arXiv:2212.11270}, 2022.

\end{thebibliography}
}

\end{document}